\documentclass[10pt,twocolumn,letterpaper]{article}

\usepackage{cvpr}              

\usepackage{colortbl}
\usepackage{pifont}
\definecolor{oursyellow}{HTML}{FCEFCB}
\definecolor{gaingreen}{HTML}{1E8449}

\newcommand{\gain}[1]{{\scriptsize\textcolor{gaingreen}{$\downarrow$#1}}}
\definecolor{citeblue}{rgb}{0.21,0.49,0.74}   
\newcommand{\venue}[2]{{\hfill \color{gray}[{\it #2}]}}

\newcommand{\reader}{\mathrm{reader}}
\newcommand{\writer}{\mathrm{writer}}

\newcommand{\trip}[3]{(\text{#1},\,\text{#2},\,\text{#3})}

\newcommand{\method}{VSG\xspace}
\newcommand{\rg}{Response-G1\xspace}
\newcommand{\mg}{visual-state graph\xspace}

\definecolor{cvprblue}{rgb}{0.21,0.49,0.74}
\usepackage[pagebackref,breaklinks,colorlinks,allcolors=cvprblue]{hyperref}

\usepackage{booktabs}
\usepackage[table]{xcolor}
\usepackage{adjustbox}
\usepackage{amssymb}
\usepackage{multirow}
\usepackage{array}
\usepackage{amsmath}

\def\confName{CVPR}
\def\confYear{2027}

\title{Visual-State Graph Memory for Streaming Video Understanding}

\author{
Ailar Mahdizadeh \quad
Pritam Sarkar \quad
Xiangteng He \quad
Leonid Sigal\\
University of British Columbia\\
Vector Institute for Artificial Intelligence\\
{\tt\small ailar.mahdizadeh@ubc.ca}
}

\begin{document}
\twocolumn[{%
\renewcommand\twocolumn[1][]{#1}%
\maketitle
\vspace{-1.6em}
\begin{center}
  \captionsetup{type=figure}
  \includegraphics[width=1\textwidth]{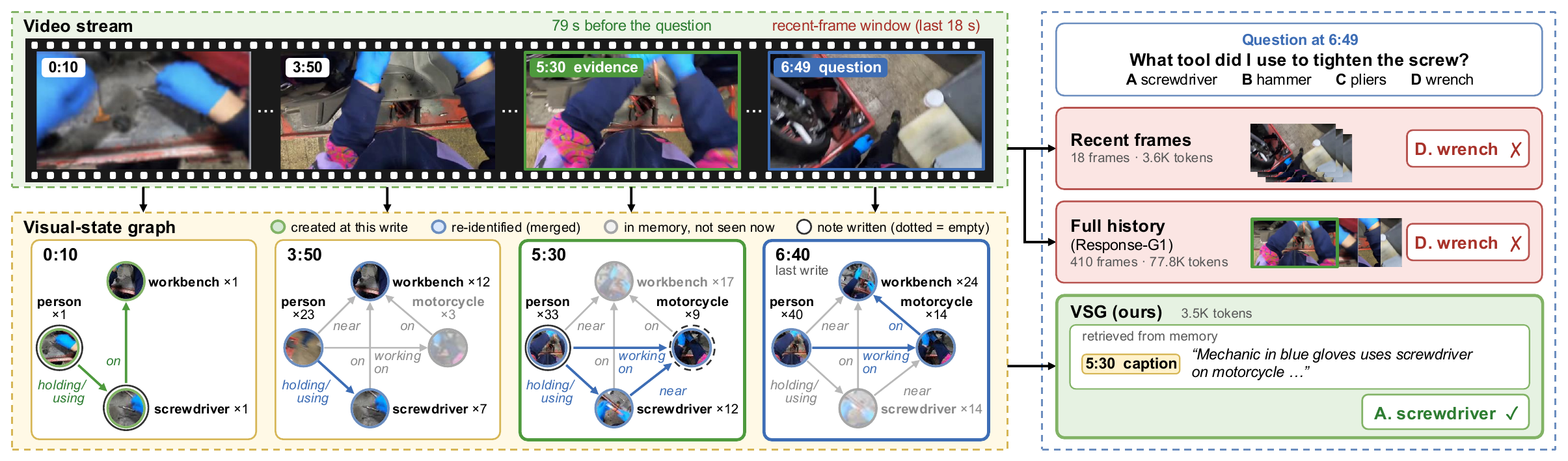}
  \vspace{-0.9em}
\caption{\textbf{Preserving frames vs.\ preserving state.} VSG merges repeated
observations of an object into one graph node ($\times n$ counts merged
observations) and stores a caption per clip. A caption from 5:30 answers the
question at 6:49. Recent frames no longer reach that moment, and Response-G1
answers wrongly despite reading the full history. VSG uses the same VLM and
token budget as recent frames.}
  \label{fig:overview}
\end{center}
\vspace{-0.2em}
}]
\begin{abstract}
In streaming video understanding, the evidence needed to answer a question may have already left the recent-frame window. Keeping all frames does not scale with video length, while dropping older frames limits the model's ability to reason about the past. We propose Visual-State Graph (VSG)\footnote{The project page is available at \url{https://ailarmhz.github.io/VSG/}.}, a training-free memory that stores the state of a stream rather than the raw sequence of frames. VSG represents the stream as a graph of objects and their relations, together with a log of events; similar to a dynamically evolving scene graph. An object observed repeatedly is represented by one persistent node, with a timestamped history of how its state changes. To determine whether a newly mentioned object already exists in the graph, we compare the relations it takes part in with those of stored objects.  We build this memory as the video arrives, before the questions are known. When a question arrives, we retrieve relevant objects and events alongside recent frames under a fixed input budget. Our evaluation spans four streaming-video benchmarks and seven open-source vision-language models ranging from 7B to 32B parameters. Under matched input budgets, VSG improves reasoning about the past while preserving real-time understanding.

\end{abstract}
    
\section{Introduction}
\label{sec:intro}

Streaming video understanding enables interactive video assistants and autonomous agents to respond to events as they unfold~\cite{lin2024streamingbench,li2025ovobench}. It requires models to interpret the current scene, recall earlier objects or events when queried, and respond when a question is phrased.
 
Despite these demands, a simple baseline that feeds only a window of recent frames to an off-the-shelf vision-language model (VLM) already matches or surpasses published streaming methods~\cite{shen2026simplestream}. A recent-frame window provides high-fidelity evidence of the present but loses access to earlier observations once they fall outside the window. In \cref{fig:overview}, a user asks which tool tightened a screw 79 seconds after the action occurred, when the relevant frames are no longer available. Keeping all frames preserves the visual history but does not scale, as the model's input grows with video duration. 

To address this limitation, existing methods retain historical information through different forms of memory. Some compress visual history into a bounded set of visual tokens or key-value (KV) cache entries~\cite{kim2025infinipotv,yang2025streammem,mahdizadeh2026cords}. Others construct structured textual memories organized as events~\cite{liang2026oasis}, scene graphs~\cite{ma2026responseg1}, or entities~\cite{fan2026folio}. However, retaining historical information does not guarantee accurate answers. In the example in \cref{fig:overview}, even a full-history scene-graph baseline~\cite{ma2026responseg1} answers incorrectly despite processing substantially more input tokens. Moreover, under a fixed input budget, textual memories and recent frames compete for the same tokens, and allocating more tokens to history can reduce accuracy on the current scene~\cite{shen2026simplestream}. The central challenge is therefore to support accurate reasoning about the past while preserving understanding of the present under a fixed input budget.

To this end, we introduce the Visual-State Graph (\method), a training-free memory that represents the state of a video stream through persistent objects and their evolving states. Instead of storing each observation of an object independently, we maintain a persistent graph node with a timestamped history of its descriptions. Relations between objects help determine whether a newly observed object already exists in the graph, allowing subsequent observations to update its history. A separate event log records actions as the video unfolds, preserving the sequence of events alongside the evolving object states.

We construct this memory online using a frozen vision-language model, which we call the \emph{writer}. As the video arrives, the writer periodically extracts objects, their relations, and a short event description to update the graph and event log. When a question arrives, we retrieve relevant object histories and events and provide them alongside recent frames to another frozen VLM, which we call the \emph{reader}. The reader operates under the same input-token budget as the recent-frame baseline. Since the memory is constructed before questions are known, it can be built once per video and reused across different reader models. For proactive tasks, we construct a separate memory conditioned on an instruction provided before the stream.

We evaluate \method across four streaming-video benchmarks and seven open-source VLMs ranging from 7B to 32B parameters. Under matched input-token budgets, \method improves aggregate performance on each benchmark, with consistent gains in backward tracing and real-time understanding on OVO-Bench \cite{li2025ovobench} and StreamingBench \cite{lin2024streamingbench}, respectively. Our ablations further demonstrate the importance of persistent object-state histories and show that historical memory complements recent visual context. 

\noindent Our main {\bf contributions} are summarized below.
\begin{itemize}
    \item {We introduce \method, a training-free memory that represents persistent object identities, their timestamped state histories, relations, and a separate event log. The memory is constructed before questions are known and can be reused across different VLMs.}
    \item {We develop relational matching to consolidate repeated object observations and budget-controlled retrieval to combine relevant historical information with recent frames.}
    \item {We evaluate \method across four streaming-video benchmarks and seven open-source VLMs under matched input-token budgets. Controlled ablations demonstrate the importance of persistent object-state histories and their complementarity with recent visual information.}
\end{itemize}
\section{Related Work}
\label{sec:related}

\paragraph{Streaming video understanding.}
Streaming video-language models enable real-time interaction with ongoing streams~\cite{videollmonline,dispider,streamagent,wu2024videollmmod,huang2025videochatonline,chen2025livecc}, including proactive and temporally aligned responses~\cite{wang2025mmduet,wang2025streambridge}.
Benchmarks such as StreamingBench~\cite{lin2024streamingbench}, OVO-Bench~\cite{li2025ovobench}, and SVBench~\cite{yang2025svbench} assess real-time perception, historical recall, and temporal reasoning.
Recent work shows that simply providing recent frames to an off-the-shelf VLM can rival specialized streaming methods~\cite{shen2026simplestream}.
We adopt this as our primary baseline under matched input-token budgets.
\paragraph{Latent video memories.}
MovieChat~\cite{song2024moviechat}, MA-LMM~\cite{he2024malmm}, Flash-VStream~\cite{flashvstream}, and TimeChat-Online~\cite{timechatonline} compress or selectively retain visual features, while StreamForest~\cite{zeng2025streamforest} organizes history into adaptive event-based memory.
Other methods manage key-value (KV) caches to reduce streaming inference costs~\cite{di2025rekv,kim2025infinipotv,yang2025streammem,mahdizadeh2026cords}.
These model-dependent representations limit cross-architecture reuse, whereas \method maintains explicit semantic memory accessible to different frozen VLMs.

\paragraph{Structured semantic memories.}
OASIS~\cite{liang2026oasis} organizes observations into hierarchical events, while \rg~\cite{ma2026responseg1} retrieves query-guided scene graphs for proactive responses.
FOLIO~\cite{fan2026folio} tracks entity states, relations, and actions, relying primarily on names and attributes for identity association.
In contrast, \method leverages relational context to distinguish similar objects and builds question-independent memory reusable across reader models.
Retrieval combines relevant object-state histories and events with recent frames under a controlled token budget.
\section{Method}
\label{sec:method}

\begin{figure*}[t]
  \centering
  \includegraphics[width=1\textwidth]{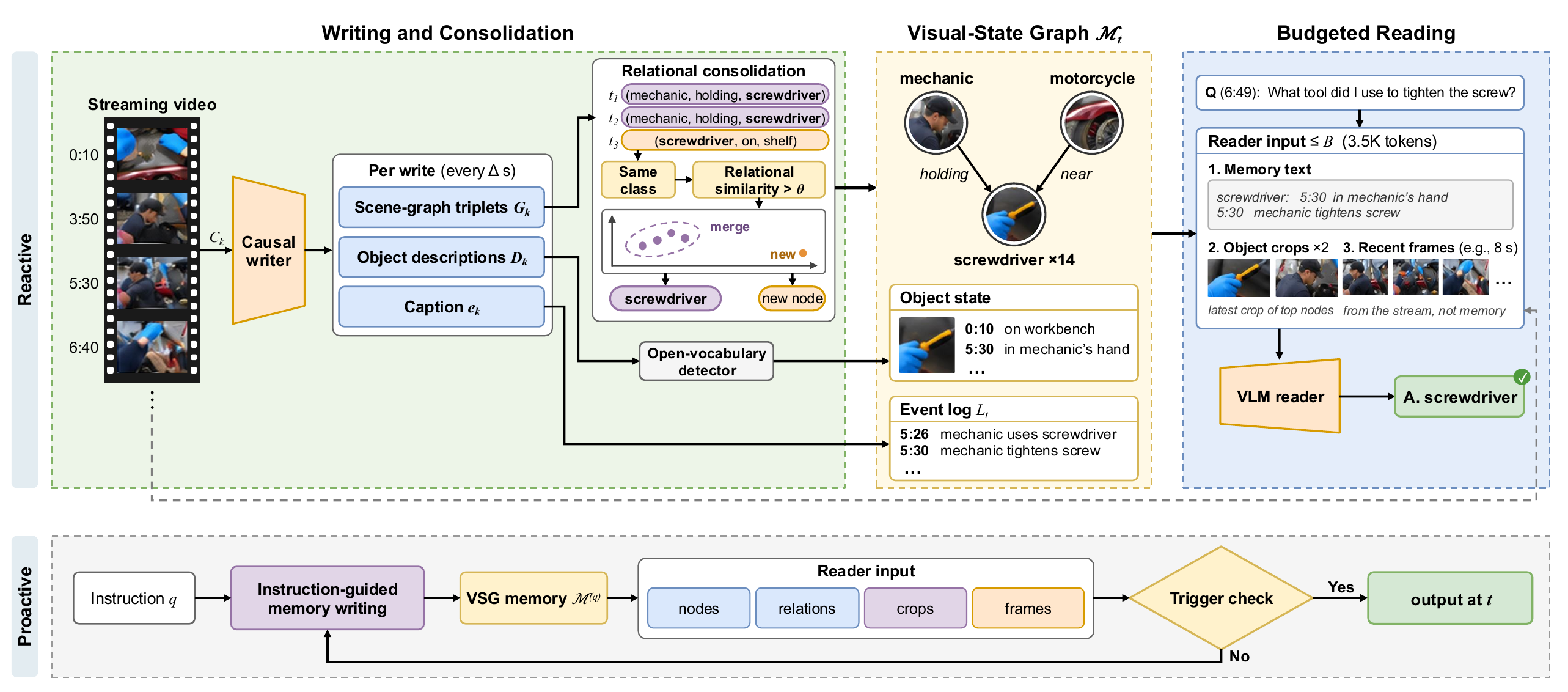}
  \caption{\textbf{Visual-State Graph (VSG).} \emph{Reactive} (top): before any
question, a frozen writer describes each clip; relational consolidation keeps one
persistent node per object, and captions form the event log. The frozen reader
answers from memory text, object crops and recent frames within budget tokens.
\emph{Proactive} (bottom): a per-instruction memory is written every second and
checked by a yes/no trigger.}
  \label{fig:method}
\end{figure*}

Our core idea is to store the evolving state of a video stream instead of treating each object observation as a separate record.
We represent each object as a persistent node with a timestamped history of its descriptions, and keep events in a separate log.
To write this memory, we use a frozen VLM, the \emph{writer}, to describe the stream as frames arrive. The writer has no access to future questions, allowing the same memory to be reused across questions. To answer a question, we give a second frozen VLM, the \emph{reader}, relevant information from the memory together with recent frames (\cref{fig:method}).
We select the retrieved information and recent frames under a fixed token budget for the reader.

\subsection{Problem setup}
\label{sec:setting}

Let $I_1,I_2,\dots$ denote the frames of a video stream sampled at one frame per second.
At decision time $T$, a frozen $\reader$ answers a
question $q$ using an input context $X_T$ of at most $B$
tokens, including the question itself. Ideally, this context would maximize the probability of the correct answer $a_T$ within a token budget $B$:
\begin{equation}
    X_T^\star =
    \operatorname*{arg\,max}_{X:\,|X|\leq B}
    \Pr_{\reader}(a_T \mid X),
    \label{eq:ideal}
\end{equation}
where $|X|$ counts the context tokens passed to the reader. We cannot compute this ideal context because the correct answer is unknown and the possible contexts include subsets of all past evidence.
A recent-frame window discards older evidence, whereas retaining the full history eventually exceeds any fixed token budget.
We therefore construct $X_T$ from recent frames $\mathcal{F}_T$ and evidence retrieved from a memory $\mathcal{M}_T$, with $|X_T|\leq B$.
For budget-matched comparisons, we set $B$ to the input-token budget of a recent-frame window containing $N$ frames. 
The time at which $q$ becomes available determines what information the $\writer$ may use during memory construction. Specifically, we allow the $\writer$ to handle both scenarios: when the stream is ongoing and the question is not yet known (i.e., arrives in the future), the $\writer$ builds the memory without access to $q$, which we refer to as the \emph{reactive} setting; alternatively, when an instruction is given prior to the stream requiring a response once a condition is met, the $\writer$ may use $q$ as an auxiliary signal to guide memory creation, referred to as the \emph{proactive} setting.

In both settings, the memory depends only on frames observed up to time $T$,
\begin{equation}
    \mathcal{M}_T =
    \begin{cases}
        \writer(I_{1:T}) & \text{(reactive)},\\
        \writer(I_{1:T};\,q) & \text{(proactive)}.
    \end{cases}
    \label{eq:modes}
\end{equation}

\subsection{The Visual-State Graph}
\label{sec:memory}

A scene graph represents an image through object nodes and their relations, often expressed as subject-predicate-object triplets ~\cite{johnson2015image,krishna2017visual}.
For example, \trip{mechanic}{holding}{screwdriver} describes a relation between a mechanic and a screwdriver.
However, a scene graph of a single frame does not preserve object identity across time.
The same screwdriver may appear in many frames, but each frame provides a separate observation.
We extend scene graphs with persistent object identities, timestamped state histories, and a separate event log.

We define the memory at time $t$ as
\begin{equation}
    \mathcal{M}_t = (V_t, E_t, \mathcal{L}_t),
    \label{eq:memory}
\end{equation}
where $V_t$ contains object nodes, $E_t$ contains their relations, and $\mathcal{L}_t$ is an event log.
Each node $v \in V_t$ represents a tracked object and stores
\begin{equation}
    v = \big(
        c_v,\,
        t_v^{\mathrm{f}},\,
        t_v^{\mathrm{l}},\,
        \mathcal{D}_v,\,
        \mathcal{O}_v
    \big),
    \label{eq:node}
\end{equation}
where $c_v$ is the object label assigned by the writer, such as ``screwdriver'', and $t_v^{\mathrm{f}}$ and $t_v^{\mathrm{l}}$ denote its first and most recent observation times. The description history $\mathcal{D}_v$ contains timestamped records $(\tau,d)$, where $\tau \leq t$ is the observation time and $d$ is a short text describing the object's \emph{state} at that time, i.e., its position or condition. The visual history $\mathcal{O}_v$ contains records $(\tau,b,x)$, where $b$ is the object's bounding box and $x$ is its image crop.
For example, a screwdriver may first lie on a workbench and later appear in a mechanic's hand.
Both observations contribute to the history of the same object node.

Each edge in $E_t$ represents a relation $(u,p,w)$, where $u,w \in V_t$ are the subject and object nodes, and $p$ is a textual relation such as ``holding''. For each relation, we store its first and most recent observation times and the number of times it was observed.
The event log $\mathcal{L}_t$ contains an ordered sequence of timestamped captions $(\tau,e)$, with one caption per writer call.
For example, an event may describe a mechanic tightening a bolt on a motorcycle. Here, \emph{state} refers to the properties and relations associated with persistent objects over time. Descriptions capture changes in position or condition, such as a screwdriver moving from a workbench to a mechanic's hand.
Image crops preserve visual details, such as scratches and texture on the screwdriver's handle, that short descriptions may omit.
The event log complements these object histories with scene-level actions and interactions.
An object's description may change, but its identity remains associated with the same node when subsequent observations match it. We store memory updates in an append-only record, which allows us to reconstruct $\mathcal{M}_t$ from updates observed up to time $t$.
The memory retains object crops and structured descriptions, but not complete past frames.
Its storage is not bounded by the reader's token budget $B$. Only the context retrieved from memory and the recent frames must fit within that budget (\cref{sec:reader}).

\subsection{Writing and consolidation}
\label{sec:writer}
\label{sec:consolidation}

\textbf{Writing.} We build the memory in a single online pass over the video.
Every $\Delta$ seconds, the frozen $\writer$ processes a short clip $C_k$ ending at time $t_k$, where $k$ indexes writer calls.
For each clip, the writer produces
\begin{equation}
    (G_k, e_k, \mathcal{D}_k) = \writer(C_k),
    \label{eq:writer}
\end{equation}

where $G_k=\{(s_i,p_i,o_i)\}$ contains scene-graph triplets, $e_k$ is a one-sentence caption, and $\mathcal{D}_k=\{(\ell_j,d_j)\}$ contains short object descriptions.
All three are generated as text by the writer in response to a fixed prompt. In each triplet, $s_i$ and $o_i$ are the subject and object labels, $p_i$ is their relation, and $i$ indexes the triplets from the current call.
Each pair $(\ell_j,d_j)$ associates an object label $\ell_j$ with its description $d_j$, and $j$ indexes the descriptions.
Note that $\mathcal{D}_k$ is the output of a single writer call, whereas $\mathcal{D}_v$ in \cref{eq:node} is the history stored by one node.
Each $d_j$ is appended, with its observation time, to $\mathcal{D}_v$ of the node $v$ that its label $\ell_j$ is matched to.
Together, these outputs update the three components of $\mathcal{M}_{t_k}=(V_{t_k},E_{t_k},\mathcal{L}_{t_k})$ in \cref{eq:memory}.
The labels in $G_k$ are matched to nodes in $V_{t_k}$, existing or newly created, by the consolidation step below, and each triplet then adds or refreshes an edge in $E_{t_k}$.
We append $(t_k,e_k)$ to the event log $\mathcal{L}_{t_k}$. We limit the number of object descriptions per call to reduce the cost of writing.
We prioritize newly created nodes that have no description, followed by nodes whose latest descriptions are the oldest.
Since triplets can create nodes independently of descriptions, a node may initially have an empty description history.
We localize each described object using the open-vocabulary detector OWLv2~\cite{minderer2023owlv2}, and store its bounding box and image crop.
The resulting reactive memory is independent of the questions and can serve multiple queries and readers without another pass through the video.
\paragraph{Relational consolidation.}
A writer may mention the same physical object in many clips.
If we create a new node for every mention, the memory accumulates repeated records of the same object.
We therefore associate each mention with an existing node or create a new node.
This is an online data-association problem~\cite{teng2021target}.

We use relational context to compare mentions.
For example, a screwdriver may appear in
\trip{mechanic}{holding}{screwdriver} at multiple times.
The shared relation provides evidence that the mentions refer to the same object.
Unlike crop appearance, relational text also describes the objects and actions surrounding a mention.
Relations can change, however, so we combine the current context with the history accumulated by each node rather than treat a single relation as a fixed identity.

For a label $\ell$ appearing in $G_k$, let $G_k(\ell)$ contain the triplets whose subject or object label is $\ell$.
We represent its relational context as
\begin{equation}
    \mathbf{r}_k(\ell)
    =
    \frac{1}{|G_k(\ell)|}
    \sum_{(s,p,o)\in G_k(\ell)}
    \phi(s,p,o),
    \label{eq:context}
\end{equation}
where $\phi(s,p,o)$ is the mean of the writer's input token embeddings for the text ``$s$ $p$ $o$''.
For each stored node $v$, we maintain a running mean $\bar{\mathbf{r}}_v$ of the relational contexts previously assigned to that node. A synonym map $\kappa$ groups equivalent object labels into the same class.
For example, ``motorbike'' and ``motorcycle'' can share a class.
Given a new label $\ell$, we consider only nodes whose labels belong to the same class,
\begin{equation}
    \mathcal{C}_k(\ell)
    =
    \{v\in V_{t_k^-}:
    \kappa(c_v)=\kappa(\ell)\},
    \label{eq:candidates}
\end{equation}
where $V_{t_k^-}$ denotes the nodes stored before the current write.
If the candidate set is nonempty, we select
\begin{equation}
    v^\star
    =
    \operatorname*{arg\,max}_{v\in\mathcal{C}_k(\ell)}
    \cos\big(
        \mathbf{r}_k(\ell),
        \bar{\mathbf{r}}_v
    \big).
\end{equation}
We then resolve the mention using
\begin{equation}
    \operatorname{resolve}_k(\ell)
    =
    \begin{cases}
        v^\star,
        &
        \begin{aligned}
        &\mathcal{C}_k(\ell)\neq\emptyset
        \text{ and}\\[-2pt]
        &\cos\big(
            \mathbf{r}_k(\ell),
            \bar{\mathbf{r}}_{v^\star}
        \big)>\theta,
        \end{aligned}\\[4pt]
        v_{\mathrm{new}},
        & \text{otherwise},
    \end{cases}
    \label{eq:resolve}
\end{equation}
where $\theta$ is the similarity threshold and $v_{\mathrm{new}}$ denotes a newly created node.
For a successful match, we update the node's running relational context, observation count, and most recent observation time.
Any new description and crop extend its history.
Otherwise, we create a node for the new mention. Consider a mechanic holding a screwdriver.
When the writer first reports \trip{mechanic}{holding}{screwdriver}, we create a screwdriver node if no existing node matches.
A later mention with a similar relational context can match this node, so we update its history instead of creating another node.
If the screwdriver later appears in a different relation, the accumulated context may still support the match, although a sufficiently large change can cause a new node to be created.
We then update the relation edges using the resolved object nodes.
For repeated relations, we increment their observation counts and refresh their most recent observation times.

\subsection{Reading under a fixed token budget}
\label{sec:reader}

When a question $q$ arrives at time $T$, the reader selects evidence from the current memory $\mathcal{M}_T$ and the recent frames.
We allocate the input-token budget $B$ in three steps, with memory text first, object crops second, and video frames last.
This priority preserves historical information before allocating the remaining tokens to visual observations.
\paragraph{Retrieving memory.}
We rank object nodes by the relevance of their latest descriptions to the question.
For each node $v$ with at least one description, we compute
\begin{equation}
    \rho_v
    =
    \cos\big(
        \psi(q),
        \psi(c_v \oplus d_v^{\mathrm{last}})
    \big),
    \label{eq:rank}
\end{equation}
where $\psi$ is a CLIP text encoder~\cite{radford2021clip}, $\oplus$ concatenates the object label and description, and $d_v^{\mathrm{last}}$ is the most recent description of node $v$.
Using the latest description allows retrieval to consider the object's current state rather than its label alone. We select the top-$k$ nodes and retain their first and most recent observation times, their first description, and their two latest descriptions.
This gives the reader both the initial state and the recent history of each selected object.
We also retrieve event lines from $\mathcal{L}_T$.

If the event log fits within the allocated text budget, we retain it in full. Otherwise, we select its most relevant lines using text similarity to $q$, together with the latest lines, and arrange them in temporal order.
We denote the serialized memory text by $\sigma_T$.

\vspace{-0.15in}

\begin{table*}[t]
  \centering
  \providecolor{oursyellow}{HTML}{FCEFCB}
  \providecolor{gaingreen}{HTML}{1E8449}
  \providecolor{citeblue}{rgb}{0.21,0.49,0.74}
  \providecommand{\method}{VSG}
  \providecommand{\rg}{Response-G1}
  \providecommand{\pub}[1]{\textcolor{gray}{#1}}
  \providecommand{\venue}[2]{{\color{gray}[\hyperlink{cite.#1}{#2}]}}
  \providecommand{\gainpt}[1]{{\tiny\textcolor{gaingreen}{(#1)}}}
  \providecommand{\cref}[1]{Sec.~\ref{#1}}
  \newcommand{\bb}[3]{\cellcolor{white}\multirow{-#1}{*}{\shortstack[l]{\textbf{#2}\\#3}}}
  \caption{\textbf{Main results} (accuracy, \%).
Each block uses a frozen reader; all \method{} rows share the same memory except for proactive evaluation (\cref{sec:proactive}), where input is not capped at $B$.
Tok./q is the mean reader input tokens per real-time/backward query; Avg.\ averages these two accuracies, and Overall averages 10 real-time categories and proactive output.
Green shows gains over recent frames; bold marks the best per block.
Gray denotes vanilla inference ($^{*}$run by us).
Cited gray rows and the top three blocks use published protocols and questions.
  }
  \label{tab:main}
  \scriptsize
  \setlength{\tabcolsep}{3pt}
  \renewcommand{\arraystretch}{0.95}
  \resizebox{\textwidth}{!}{%
  \begin{tabular}{l l r c c c r c c c}
    \toprule
    & & \multicolumn{4}{c}{\textbf{OVO-Bench} \cite{li2025ovobench}} & \multicolumn{4}{c}{\textbf{StreamingBench} \cite{lin2024streamingbench}}\\
    \cmidrule(lr){3-6}\cmidrule(lr){7-10}
    Backbone & Method & Tok./q & Real-time & Backward & Avg. & Tok./q & Real-time & Proactive & Overall\\
\midrule
\multirow{2}{*}{\shortstack[l]{\textbf{Proprietary}}}
& Gemini 1.5 Pro~\cite{gemini15} & -- & 69.3 & 62.5 & 65.9 & -- & 75.7 & 45.1 & -- \\
& GPT-4o~\cite{gpt4o} & -- & 64.5 & 60.8 & 62.6 & -- & 73.3 & 56.9 & -- \\
\cmidrule(lr){1-10}
\multirow{4}{*}{\shortstack[l]{\textbf{Open-source}\\offline}}
& LLaVA-OneVision-7B~\cite{llavaonevision} \hskip 0pt plus 1filll\venue{llavaonevision}{TMLR 2025} & -- & 64.0 & 43.7 & 53.9 & -- & 71.1 & 29.6 & -- \\
& Qwen2-VL-7B~\cite{qwen2vl} & -- & 56.0 & 46.5 & 51.2 & -- & 69.0 & 22.7 & -- \\
& Qwen2-VL-72B~\cite{qwen2vl} & -- & 61.9 & 57.0 & 59.4 & -- & -- & -- & -- \\
& InternVL-V2-8B~\cite{internvl2} & -- & 60.4 & 43.4 & 51.9 & -- & 63.7 & 40.9 & -- \\
\cmidrule(lr){1-10}
\multirow{5}{*}{\shortstack[l]{\textbf{Open-source}\\streaming}}
& VideoLLM-online-8B~\cite{videollmonline} \hskip 0pt plus 1filll\venue{videollmonline}{CVPR 2024} & -- & 20.8 & 17.7 & 19.3 & -- & 36.0 & 3.9 & -- \\
& Flash-VStream-7B~\cite{flashvstream} \hskip 0pt plus 1filll\venue{flashvstream}{ICCV 2025} & -- & 28.4 & 27.4 & 27.9 & -- & 23.2 & 2.0 & -- \\
& Dispider-7B~\cite{dispider} \hskip 0pt plus 1filll\venue{dispider}{CVPR 2025} & -- & 54.5 & 36.1 & 45.3 & -- & 67.6 & 25.3 & -- \\
& TimeChat-Online-7B~\cite{timechatonline} \hskip 0pt plus 1filll\venue{timechatonline}{MM 2025} & -- & 58.6 & 42.0 & 50.3 & -- & 75.4 & 28.8 & -- \\
& StreamAgent-7B~\cite{streamagent} & -- & 61.3 & 41.7 & 51.5 & -- & 74.3 & 28.9 & -- \\
\midrule
& \pub{Vanilla}~\cite{streamingeval} & \pub{--} & \pub{74.3} & \pub{40.9} & \pub{57.6} & \pub{--} & \pub{78.0} & \pub{--} & \pub{--} \\
& Recent frames & 3.4K & 73.9 & 48.6 & 61.3 & 3.3K & 79.4 & 36.8 & 74.1 \\
& + \rg{}~\cite{ma2026responseg1} \hskip 0pt plus 1filll\venue{ma2026responseg1}{ACL 2026} & 29.9K & 69.3 & 44.9 & 57.1 & 30.9K & 75.7 & 30.0 & 71.2 \\
\rowcolor{oursyellow}
\bb{4}{InternVL3.5-8B}{\cite{internvl35}} & \textbf{+ \method{} (ours)} & 3.3K & \textbf{75.2}\,\gainpt{+1.3} & \textbf{57.5}\,\gainpt{+8.9} & \textbf{66.4}\,\gainpt{+5.1} & 3.3K & \textbf{80.0}\,\gainpt{+0.6} & \textbf{37.4}\,\gainpt{+0.6} & \textbf{74.7}\,\gainpt{+0.6} \\
\midrule
& \pub{Vanilla}~\cite{li2025ovobench} & \pub{--} & \pub{63.5} & \pub{40.4} & \pub{52.0} & \pub{--} & \pub{--} & \pub{--} & \pub{--} \\
& Recent frames & 3.2K & 72.6 & 41.7 & 57.2 & 3.1K & 78.4 & 19.5 & 71.8 \\
& + \rg{}~\cite{ma2026responseg1} \hskip 0pt plus 1filll\venue{ma2026responseg1}{ACL 2026} & 23.9K & 68.6 & 38.8 & 53.7 & 24.4K & 75.8 & 25.3 & 71.0 \\
\rowcolor{oursyellow}
\bb{4}{LLaVA-Video-7B}{\cite{llavavideo}} & \textbf{+ \method{} (ours)} & 3.3K & \textbf{73.5}\,\gainpt{+0.9} & \textbf{48.7}\,\gainpt{+7.0} & \textbf{61.1}\,\gainpt{+3.9} & 3.2K & \textbf{79.1}\,\gainpt{+0.7} & \textbf{26.8}\,\gainpt{+7.3} & \textbf{73.2}\,\gainpt{+1.4} \\
\midrule
& \pub{Vanilla}~\cite{streamingeval} & \pub{--} & \pub{75.5} & \pub{45.8} & \pub{60.7} & \pub{--} & \pub{76.2} & \pub{--} & \pub{--} \\
& Recent frames & 3.3K & 74.4 & 53.5 & 64.0 & 3.3K & 77.7 & 34.7 & 72.4 \\
\rowcolor{oursyellow}
\bb{3}{LLaVA-OV-1.5-8B}{\cite{llavaov15}} & \textbf{+ \method{} (ours)} & 3.3K & \textbf{76.9}\,\gainpt{+2.5} & \textbf{63.5}\,\gainpt{+10.0} & \textbf{70.2}\,\gainpt{+6.2} & 3.3K & \textbf{78.6}\,\gainpt{+0.9} & \textbf{36.8}\,\gainpt{+2.1} & \textbf{73.7}\,\gainpt{+1.3} \\
\midrule
& \pub{Vanilla}~\cite{liang2026oasis} & \pub{--} & \pub{66.8} & \pub{51.2} & \pub{59.0} & \pub{--} & \pub{72.8} & \pub{--} & \pub{--} \\
& Recent frames & 3.3K & 78.0 & 53.9 & 66.0 & 3.2K & 79.5 & 39.5 & 74.5 \\
& + \rg{}~\cite{ma2026responseg1} \hskip 0pt plus 1filll\venue{ma2026responseg1}{ACL 2026} & 37.5K & 73.6 & 52.1 & 62.9 & 29.4K & 77.5 & 45.8 & 73.9 \\
& + OASIS~\cite{liang2026oasis} \hskip 0pt plus 1filll\venue{liang2026oasis}{CVPR 2026} & 10.5K & \textbf{79.9} & 59.3 & 69.6 & 9.5K & 80.2 & -- & -- \\
\rowcolor{oursyellow}
\bb{5}{Qwen3-VL-8B}{\cite{qwen3vl}} & \textbf{+ \method{} (ours)} & 3.3K & 79.0\,\gainpt{+1.0} & \textbf{61.2}\,\gainpt{+7.3} & \textbf{70.1}\,\gainpt{+4.1} & 3.2K & \textbf{83.1}\,\gainpt{+3.6} & \textbf{46.3}\,\gainpt{+6.8} & \textbf{78.5}\,\gainpt{+4.0} \\
\midrule
& \pub{Vanilla\smash{$^{*}$}} & \pub{--} & \pub{74.0} & \pub{61.5} & \pub{67.8} & \pub{10.1K} & \pub{77.3} & \pub{--} & \pub{--} \\
& Recent frames & 3.3K & 77.4 & 65.2 & 71.3 & 3.2K & 82.6 & 44.2 & 78.2 \\
& + \rg{}~\cite{ma2026responseg1} \hskip 0pt plus 1filll\venue{ma2026responseg1}{ACL 2026} & 37.5K & 79.7 & 61.4 & 70.6 & 29.4K & 82.4 & 47.9 & 78.6 \\
& + OASIS~\cite{liang2026oasis} \hskip 0pt plus 1filll\venue{liang2026oasis}{CVPR 2026} & 15.3K & \textbf{82.0} & 62.9 & 72.5 & 13.2K & 82.9 & -- & -- \\
\rowcolor{oursyellow}
\bb{5}{Qwen3-VL-32B}{\cite{qwen3vl}} & \textbf{+ \method{} (ours)} & 3.3K & 79.3\,\gainpt{+1.9} & \textbf{70.2}\,\gainpt{+5.0} & \textbf{74.8}\,\gainpt{+3.5} & 3.2K & \textbf{83.6}\,\gainpt{+1.0} & \textbf{48.4}\,\gainpt{+4.2} & \textbf{79.7}\,\gainpt{+1.5} \\
\midrule
& \pub{Vanilla}~\cite{streamflow} & \pub{--} & \pub{--} & \pub{--} & \pub{--} & \pub{--} & \pub{76.3} & \pub{--} & \pub{--} \\
& Recent frames & 3.3K & 79.8 & 49.2 & 64.5 & 3.2K & 84.3 & 35.0 & 79.0 \\
& + \rg{}~\cite{ma2026responseg1} \hskip 0pt plus 1filll\venue{ma2026responseg1}{ACL 2026} & 37.6K & 77.6 & 49.6 & 63.6 & 29.4K & 84.2 & 43.5 & 79.7 \\
& + OASIS~\cite{liang2026oasis} \hskip 0pt plus 1filll\venue{liang2026oasis}{CVPR 2026} & 17.2K & 77.2 & 50.9 & 64.1 & 17.1K & 79.5 & -- & -- \\
\rowcolor{oursyellow}
\bb{5}{Qwen3.5-9B}{\cite{qwen35}} & \textbf{+ \method{} (ours)} & 3.3K & \textbf{80.2}\,\gainpt{+0.4} & \textbf{58.0}\,\gainpt{+8.8} & \textbf{69.1}\,\gainpt{+4.6} & 3.2K & \textbf{85.0}\,\gainpt{+0.7} & \textbf{47.0}\,\gainpt{+12.0} & \textbf{80.7}\,\gainpt{+1.7} \\
\midrule
& \pub{Vanilla\smash{$^{*}$}} & \pub{--} & \pub{74.5} & \pub{58.0} & \pub{66.3} & \pub{10.1K} & \pub{83.1} & \pub{--} & \pub{--} \\
& Recent frames & 3.3K & 80.8 & 58.7 & 69.8 & 3.2K & 85.9 & 43.0 & 81.1 \\
& + \rg{}~\cite{ma2026responseg1} \hskip 0pt plus 1filll\venue{ma2026responseg1}{ACL 2026} & 37.5K & 83.4 & 59.2 & 71.3 & 29.4K & 85.7 & 38.9 & 80.8 \\
& + OASIS~\cite{liang2026oasis} \hskip 0pt plus 1filll\venue{liang2026oasis}{CVPR 2026} & 13.1K & 81.7 & 65.7 & 73.7 & 10.8K & 84.0 & -- & -- \\
\rowcolor{oursyellow}
\bb{5}{Qwen3.5-27B}{\cite{qwen35}} & \textbf{+ \method{} (ours)} & 3.3K & \textbf{83.9}\,\gainpt{+3.1} & \textbf{67.9}\,\gainpt{+9.2} & \textbf{75.9}\,\gainpt{+6.1} & 3.2K & \textbf{86.7}\,\gainpt{+0.8} & \textbf{54.7}\,\gainpt{+11.7} & \textbf{83.1}\,\gainpt{+2.0} \\
    \bottomrule
  \end{tabular}}
\end{table*}

\paragraph{Allocating visual tokens.}
After selecting the memory text, we include the most recent available image region of each highest-ranked object, extracted using its bounding box.
Let $n_c$ denote the number of selected crops, and let $\kappa_c$ and $\kappa_f$ denote the estimated token counts of one crop and one video frame, respectively.
We allocate the remaining budget to
\begin{equation}
    n_f =
    \max\!\left(
        0,\,
        \min\!\left(
            n_{\max},
            \left\lfloor
            \frac{
                B-|\sigma_T|-|q|-n_c\kappa_c
            }{\kappa_f}
            \right\rfloor
        \right)
    \right),
    \label{eq:nframes}
\end{equation}
where $|\cdot|$ counts text tokens and $n_{\max}$ limits the number of selected frames.
We restrict the memory text and cropped object images to fit within $B$ before allocating frames. We select up to $\min(W,n_f)$ frames from the most recent consecutive seconds, where $W$ is the maximum size of the recent-frame window. If additional slots remain, we select historical key frames from the times at which the retrieved nodes were last observed or described.
We remove duplicate frames and arrange the selected frames chronologically. The reader receives the selected crops, followed by the frames and one text turn containing $\sigma_T$ and the question.
The estimated token allocation is
\begin{equation}
    \widehat{b}_T
    =
    |\sigma_T|+|q|
    +n_c\kappa_c+n_f\kappa_f
    \leq B.
    \label{eq:budget}
\end{equation}
Because image token counts depend on the reader backbone, we verify the token count after constructing the input and remove excess frames if necessary.
The resulting reader input satisfies $|X_T|\leq B$.

We retain relation edges in the graph for consolidation, but do not serialize them directly for reactive questions.
Their role is to maintain object identities across observations.
The reader instead receives the selected object histories, event lines, crops, and frames.
The budget bounds the reader's input independently of the stream duration, although the stored memory and retrieval workload may continue to grow.

\begin{table*}[t]
  \centering
  \providecolor{oursyellow}{HTML}{FCEFCB}
  \providecolor{citeblue}{rgb}{0.21,0.49,0.74}
  \providecommand{\method}{VSG}
  \providecommand{\pub}[1]{\textcolor{gray}{#1}}
  \providecommand{\venue}[2]{{\color{gray}[\hyperlink{cite.#1}{#2}]}}
  \providecommand{\bb}[3]{\cellcolor{white}\multirow{-#1}{*}{\shortstack[l]{\textbf{#2}\\#3}}}
  \caption{\textbf{Results on OmniPro and InteractionBench.}
  Each block uses one frozen reader, named in the left column.
  Gray rows are vanilla inference as reported by each benchmark.
  We mark the best result of each block in bold.}
  \label{tab:further}
  \scriptsize
  \setlength{\tabcolsep}{3pt}
  \renewcommand{\arraystretch}{1.05}
  \resizebox{\textwidth}{!}{%
  \begin{tabular}{l l ccccccccc c cccc}
    \toprule
    & & \multicolumn{10}{c}{\textbf{OmniPro}~\cite{omnipro}} & \multicolumn{4}{c}{\textbf{InteractionBench}~\cite{interactionbench}}\\
    \cmidrule(lr){3-12}\cmidrule(lr){13-16}
    & & Event & Target & State & Snap. & Cond. & Cum. & Event & Dedup. & Step & & & & & \\
    Backbone & Method & Alert & Ground & Monitor & Count & Alert & Count & Narr. & Count & Inst. & Mean & Overall & Acc. & TA & SC\\
    \midrule
& \pub{Vanilla}~\cite{omnipro,interactionbench}
  & \pub{7.5} & \pub{2.8} & \pub{18.2} & \pub{13.1} & \pub{9.0} & \pub{\textbf{11.2}} & \pub{55.8} & \pub{31.8} & \pub{25.8}
  & \pub{19.5} & \pub{34.8} & \pub{54.2} & \pub{77.8} & \pub{10.9}\\
& + Response-G1~\cite{ma2026responseg1} \hskip 0pt plus 1filll\venue{ma2026responseg1}{ACL 2026}
  & 8.3 & 6.8 & 27.8 & 2.5 & 6.0 & 9.0 & 50.3 & 30.0 & 32.8
  & 19.3 & 42.4 & 56.7 & 72.8 & 13.1\\
& + OASIS~\cite{liang2026oasis} \hskip 0pt plus 1filll\venue{liang2026oasis}{CVPR 2026}
  & 5.0 & \textbf{9.1} & 2.8 & 7.5 & 4.5 & 5.5 & 45.2 & 31.8 & 23.0
  & 14.9 & 46.0 & 59.9 & 77.5 & 20.2\\
\rowcolor{oursyellow}
\bb{4}{Qwen3-VL-8B}{\cite{qwen3vl}} & \textbf{+ \method{} (ours)}
  & \textbf{11.6} & 8.6 & \textbf{29.7} & \textbf{16.7} & \textbf{11.8} & 10.6 & \textbf{60.5} & \textbf{35.1} & \textbf{37.9}
  & \textbf{24.7} & \textbf{49.8} & \textbf{65.2} & \textbf{87.1} & \textbf{21.4}\\
    \midrule
& \pub{Vanilla}~\cite{omnipro}
  & \pub{4.8} & \pub{2.4} & \pub{7.2} & \pub{6.0} & \pub{\textbf{9.3}} & \pub{\textbf{5.3}} & \pub{33.0} & \pub{21.3} & \pub{20.0}
  & \pub{12.1} & \pub{24.6} & \pub{27.7} & \pub{34.5} & \pub{29.5}\\
& + Response-G1~\cite{ma2026responseg1} \hskip 0pt plus 1filll\venue{ma2026responseg1}{ACL 2026}
  & \textbf{5.0} & 3.0 & 13.3 & 2.5 & 6.0 & 4.5 & 39.7 & 23.4 & 21.7
  & 13.2 & 27.3 & 35.6 & 46.1 & 31.9\\
& + OASIS~\cite{liang2026oasis} \hskip 0pt plus 1filll\venue{liang2026oasis}{CVPR 2026}
  & 3.3 & \textbf{4.5} & 8.0 & \textbf{12.5} & 6.0 & 1.5 & 45.8 & 7.4 & 26.0
  & 12.8 & 31.4 & 39.5 & 39.4 & 33.0\\
\rowcolor{oursyellow}
\bb{4}{InternVL3.5-8B}{\cite{internvl35}} & \textbf{+ \method{} (ours)}
  & 4.6 & \textbf{4.5} & \textbf{20.4} & 7.0 & 8.1 & 4.7 & \textbf{54.2} & \textbf{32.4} & \textbf{33.0}
  & \textbf{18.8} & \textbf{36.2} & \textbf{42.8} & \textbf{48.3} & \textbf{34.4}\\
    \bottomrule
  \end{tabular}}
\end{table*}

\subsection{Query-conditioned memory for proactive tasks}
\label{sec:proactive}
\label{sec:mode}

In proactive tasks, an instruction $q$ is available before the video begins and specifies a condition that should trigger a response.
We therefore construct a separate memory $\mathcal{M}_t^{(q)}$ for each instruction, as defined in \cref{eq:modes}.
The frozen writer $\writer$ processes the stream every second with $q$ included in its prompt.
It uses the same scene-graph extraction, consolidation, and object detection procedures as in the reactive setting, but prioritizes descriptions of objects relevant to the instruction.
For each node, we retain its first and most recent grounded observations. At each second, the reader receives the most relevant nodes, their relations from recent observations, up to two cropped object images, and recent frames.
We also provide a short summary of recent relation changes, including relations that newly appeared or recurred.
The reader first determines whether the condition specified by $q$ has been satisfied.
The first positive decision triggers a second reader call to generate the instructed response. The benchmark protocol determines which memory mode applies.
Reactive questions use a question-agnostic memory that can serve multiple questions and readers.
Proactive instructions use separate, instruction-conditioned memories.
Both modes use the same frozen models without additional training.
Unlike the reactive setting, proactive inference does not enforce the same fixed reader-input budget $B$.

\section{Experiments}
\label{sec:experiments}

\begin{figure*}[t]
  \centering
  \includegraphics[width=0.85\textwidth]{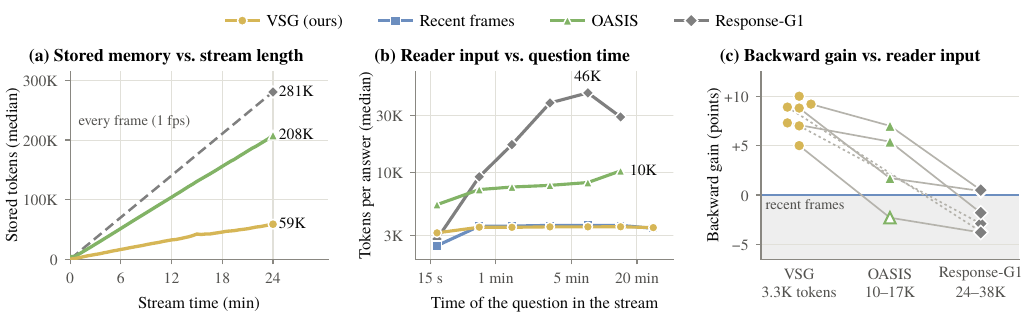}
  \caption{\textbf{Memory Gains and Costs.}
(a)~Stored memory against stream length; the dashed line keeps every frame at 1\,fps.
(b)~Reader input against question time.
(c)~Backward-tracing gain over recent frames per reader in \cref{tab:main}, with methods ordered by reader input.}\vspace{-0.15in}
  \label{fig:cost}
\end{figure*}

\subsection{Experimental setup}
\label{sec:experiment_setup}

\paragraph{Benchmarks.}
We evaluate \method{} on four streaming benchmarks.
OVO-Bench~\cite{li2025ovobench} covers real-time perception of the current scene and \emph{backward tracing}, which asks about earlier parts of the stream and directly probes memory.
StreamingBench~\cite{lin2024streamingbench} covers 10 real-time categories and a proactive output task, in which the model must respond unprompted once a specified condition occurs.
OmniPro~\cite{omnipro} and InteractionBench~\cite{interactionbench} are dedicated to proactive behavior and test whether a system speaks at the right moment and stays silent otherwise.
Further details on each benchmark are provided in \cref{supp:benchmarks} of the supplementary material.
\vspace{-0.15in}

\paragraph{Models and baselines.}
We use seven frozen VLM readers from three families (\cref{tab:main}), with Qwen3-VL-8B as the writer and OWLv2~\cite{minderer2023owlv2} for localization.
Every baseline uses the same reader as \method{}.
\emph{Vanilla} feeds the reader its standard video input.
\emph{Recent frames} fills the same token budget as \method{} with the latest frames, so it isolates the effect of our memory.
\rg~\cite{ma2026responseg1} and OASIS~\cite{liang2026oasis} are memory methods that read 3 to 11$\times$ more tokens than \method{}.
Since the methods differ in input size, we report the reader input tokens per question next to each accuracy, so that accuracy can be compared at its cost. The full pipeline and all settings are in the supplementary material, \cref{supp:impl}.

\subsection{Main results}
\label{sec:main_results}

\textbf{\method{} remembers, at a fraction of the context.} \Cref{tab:main} shows that \method{} consistently improves backward tracing without compromising real-time understanding. These improvements hold across seven VLM readers from three model families (InternVL, LLaVA, and Qwen), without reader-specific training. With the same input budget as recent frames, backward tracing improves by 16\% on average and up to 19\% in relative terms, and real-time accuracy never decreases. The proactive output (PO) task of StreamingBench, where input is not capped, improves by up to 37\%. The gains also hold as readers scale from 8B to 32B parameters. Backward tracing improves by 14\% and 8\% for Qwen3-VL-8B and 32B, and by 18\% and 16\% for Qwen3.5-9B and 27B, so larger readers still benefit from the memory. More context does not give the same benefit (\cref{fig:cost}). \method{} outperforms \rg and OASIS on backward tracing for every shared reader while reading 3 to 11$\times$ fewer tokens, and \rg falls below recent frames for four of its six readers despite reading the full history. Scores for real-time and backward tasks are in \cref{supp:pertask} of the supplementary material.

\vspace{-0.1in}
\paragraph{Proactive benchmarks.} StreamingBench includes a single proactive task, whereas OmniPro and InteractionBench (\cref{tab:further}) are dedicated to deciding \emph{when} to respond. The model must speak unprompted once a condition is met and stay silent otherwise. \method{} achieves the best OmniPro mean with both readers, with relative gains of 27\% and 55\% over vanilla inference, including large gains on tasks that depend on past state, such as State Monitoring (1.6$\times$ and 2.8$\times$). On InteractionBench, \method{} improves the overall score by 43\% and 47\% and achieves the best accuracy, timing accuracy, and silence compliance with both readers.

\subsection{Ablation Studies}
\label{sec:ablation}

We ablate historical memory, persistent object identities, and state histories (Tab.~\ref{tab:ablation}).
Compared with recent frames alone, \method{} substantially improves backward tracing while preserving real-time accuracy.
Conversely, removing recent frames leads to a considerable decline in real-time performance, highlighting the complementary roles of historical memory and current visual evidence.
Removing object nodes or replacing consolidation with a flat list reduces backward-tracing accuracy, suggesting that structured object representations help preserve historical evidence.
The remaining rows, together with appearance-only matching and swapped object histories, which test relational association and consistent object-state histories, are discussed in Sec.~\ref{sec:supp_ablation} of the supplementary material.

\begin{table}[t]
  \centering
  \caption{
    \textbf{Ablation study} with Qwen3-VL-8B.
    Accuracy (\%).}
  \label{tab:ablation}
  \scriptsize
  \setlength{\tabcolsep}{3.5pt}
  \renewcommand{\arraystretch}{1}

  \begin{tabular}{lccc}
    \toprule
    & \textbf{StreamingBench}
    & \multicolumn{2}{c}{\textbf{OVO-Bench}} \\
    \cmidrule(lr){3-4}
    \textbf{Configuration} & RT & RT & Backward \\
    \midrule

    \multicolumn{4}{l}{\emph{Controlled ablations}} \\

    Recent frames only
      & 79.5 & 78.0 & 53.9 \\

    Memory only (no recent frames)
      & 68.1 & 62.6
      & 54.8 \\

    No object nodes
      & 79.8 & 77.8 & 57.5 \\

    No consolidation (flat list)
      & 79.5 & 75.6 & 59.5 \\

    Label-only matching
      & 78.4 & 77.5 & 59.8 \\

    Timestamps shuffled
      & 79.0 & 76.3 & 59.0 \\

    Latest description only
      & 78.8 & 77.4 & 60.6 \\

    \textbf{Full \method}
      & \textbf{83.1} & \textbf{79.0}
      & \textbf{61.2} \\

    \bottomrule
  \end{tabular}
\end{table}

\begin{table}[t]
\centering
\caption{\textbf{Efficiency with Qwen3-VL-8B} on the OVO-Bench and StreamingBench questions answered by all methods. \textbf{Bold} is best, and \textcolor{gaingreen}{green} is our gain over the next best.}
\label{tab:efficiency}
\setlength{\tabcolsep}{3pt}
\renewcommand{\arraystretch}{1.1}
\resizebox{\linewidth}{!}{%
\begin{tabular}{lccccr}
\toprule
 & \multicolumn{3}{c}{Cost per question} & Memory & Latency \\
\cmidrule(lr){2-4}
Method & Tokens (mean / P95) & KV (GB) & Calls/q & (24\,min) & (s) \\
\midrule
Recent Frames & 3.4K / 3.7K & 0.51 & \textbf{1} & -- & 5.2 \\
Response-G1 \venue{ma2026responseg1}{ACL 2026} & 32.9K / 73.2K & 4.78 & 2 & 281K & 20.8 \\
OASIS \venue{liang2026oasis}{CVPR 2026} & 10.5K / 12.3K & 1.23 & 1.28 & 208K & $\geq$8.0 \\
\midrule
\rowcolor{oursyellow}
\textbf{VSG (ours)} & \textbf{3.4K / 3.6K} & \textbf{0.50} & \textbf{1}
  & \textbf{59K} \gain{3.5$\times$} & \textbf{1.0} \gain{5.2$\times$} \\
\bottomrule
\end{tabular}}
\vspace{-0.2in}
\end{table}

\subsection{Efficiency}
\label{sec:efficiency}
We measure each method's cost with Qwen3-VL-8B as the reader on a single L40S GPU. \method{} and recent frames run on HF Transformers, OASIS runs on its official HF implementation, and \rg runs on vLLM. All results use the OVO-Bench and StreamingBench questions answered by all methods. Memory is the median number of tokens a method has stored after 24 minutes of stream, as opposed to the tokens it reads per question. \Cref{tab:efficiency} and \cref{fig:cost} show that \method{} is the only method that keeps the past at the per-question cost of recent frames. It answers each question in a single reader call over 3.4K input tokens, with the same KV-cache memory as recent frames. Compared with \rg and OASIS, \method{} uses 9.7$\times$ and 3.1$\times$ fewer input tokens and has 9.6$\times$ and 2.5$\times$ lower KV-cache memory. Its stored memory after 24 minutes is 4.8$\times$ lower than keeping every frame and 3.5$\times$ lower than OASIS. \method{} has the lowest latency at 1.0\,s per question, which is 5.2$\times$ lower than recent frames, at least 8$\times$ lower than OASIS and 20.8$\times$ lower than \rg. Its reader time matches that of recent frames, and the gain comes from decoding fewer frames. \rg runs on vLLM, which is generally faster than HF Transformers, so this comparison favors \rg.

\begin{figure}[t]
  \centering
  \includegraphics[width=0.7\linewidth]{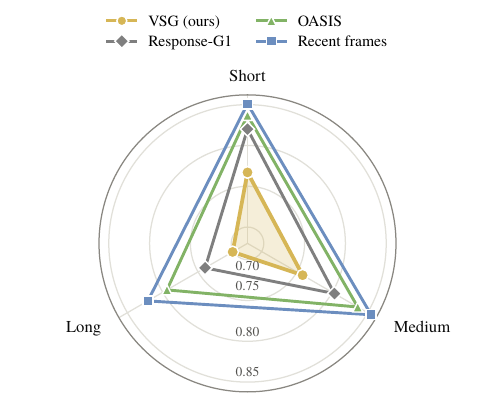}
  \caption{\textbf{Visual redundancy across stream durations.} Mean pairwise CLIP similarity of reader images (lower is better).\vspace{-0.15in}}
  \label{fig:redundancy}
\end{figure}

\subsection{Evidence redundancy analysis}
\label{sec:redundancy}
\Cref{fig:redundancy} reports the mean pairwise CLIP similarity among the images each method passes to the reader, for questions asked at short (under 2\,min), medium (2 to 10\,min), and long (over 10\,min) stream durations. Recent frames are the most redundant in all three groups, since consecutive frames show nearly the same scene. \method{} passes the least redundant images in every group, and its margin is largest on long streams, where memory matters most. The budget is thus spent on distinct evidence rather than on repeated views. \Cref{fig:overview} illustrates one such case, where the answer lies in an event recorded 79\,s before the question. \method{} retrieves it, while recent frames and \rg both answer wrongly. Additional qualitative  examples are in \cref{supp:qual} of the supplementary material.

\section{Conclusion}
\label{sec:conclusion}
We presented Visual-State Graph, a training-free memory for streaming video that stores the state of a stream rather than its sequence of frames. VSG keeps one persistent node per object with a timestamped history of how its state changes, uses the relations an object takes part in to decide whether a new mention refers to a stored object, and records events in a separate log. The memory is built as the video arrives, before the questions are known, and can be reused across frozen readers. On OVO-Bench and StreamingBench, under matched input budgets, VSG improves reasoning about the past for all seven VLMs while preserving real-time understanding, and it also improves overall performance on OmniPro and InteractionBench. Our ablations show that the memory complements recent frames. The reader's input is bounded, but the stored memory grows with the stream, and reducing the cost of writing and storing it is left to future work.
\clearpage
\section*{LLM Usage Statement}
Any LLM use was limited to writing assistance and does not affect the scientific content. All text produced by the LLM was thoroughly checked by the authors.
{\small\bibliographystyle{ieeenat_fullname}\bibliography{main}}


\ifdefined\SUPPSTANDALONE\else
  \backrefsetup{disable}
  \clearpage
  \hypersetup{pageanchor=false}
\fi
\providecommand{\mref}[1]{\cref*{#1}}
\providecommand{\Mref}[1]{\Cref*{#1}}

\setcounter{page}{1}
\maketitlesupplementary
\appendix

\setcounter{figure}{0}\setcounter{table}{0}\setcounter{equation}{0}
\renewcommand{\thefigure}{S\arabic{figure}}
\renewcommand{\thetable}{S\arabic{table}}
\renewcommand{\theequation}{S\arabic{equation}}
\renewcommand{\theHfigure}{supp.\arabic{figure}}
\renewcommand{\theHtable}{supp.\arabic{table}}
\renewcommand{\theHequation}{supp.\arabic{equation}}

\noindent Sections, figures and tables of this supplement carry a letter or
an ``S'' (\cref{supp:impl}, \cref{tab:hparams}). Numbers without one, such as
\mref{tab:main} or \mref{sec:method}, refer to the main paper.

\section{Implementation details}
\label{supp:impl}

This section details the pipeline of \method using the notation of
\mref{sec:method}. \Cref{tab:hparams} lists the settings of the two modes.
The reactive read configuration was selected on the development split
(\cref{sec:dev_ablation}), not on the held-out test questions.

\paragraph{Models and decoding.}
The writer is Qwen3-VL-8B-Instruct~\cite{qwen3vl} with greedy decoding. The
detector is OWLv2-base-patch16-ensemble~\cite{minderer2023owlv2}, and
retrieval in the reactive mode uses the text encoder of CLIP
ViT-L/14~\cite{radford2021clip}. Each of the seven readers receives its
images under a $448{\times}448$ pixel budget and decodes greedily with at
most 128 new tokens, without a system prompt. No model is fine-tuned.

\paragraph{Writing in the reactive mode.}
The stream is sampled at 1\,fps. Every $\Delta$ seconds, the writer receives
the clip $C_k$ of the last five frames and returns the triplets $G_k$, the
caption $e_k$ and the descriptions $\mathcal{D}_k$ of \mref{eq:writer}. It
sees neither a question nor a future frame. Each call describes at most
three objects, in at most 20 words each. Nodes without a description come
first, followed by the nodes whose latest description is the oldest, and a
node is described again only after at least 10\,s. Labels are grouped by the
synonym map $\kappa$, and every mention is resolved with \mref{eq:resolve}
at $\theta{=}0.55$. OWLv2 localizes each described object with a score
threshold of 0.15, and we store a 224-pixel crop of its box with 50\%
context. All updates are appended to a log, so $\mathcal{M}_T$ is rebuilt
from the writes made up to time $T$. The benchmark configuration uses
$\Delta=5$\,s for videos of at most 300\,s and $\Delta=10$\,s for longer
videos. The schedule is selected using the benchmark video duration; each write uses only the observed clip.

\paragraph{Association and state updates.}
For each label $\ell$ in $G_k$, we average the writer input-token embeddings
within each incident triplet and then average over the incident triplets
(\mref{eq:context}). Candidate nodes are restricted to the same synonym
class $\kappa(\ell)$ as in \mref{eq:candidates}. The best candidate is
accepted only when its cosine similarity exceeds 0.55; otherwise a new
node is created. If $m_v$ counts the relational contexts previously assigned
to a matched node, its running mean becomes
$\bar{\mathbf r}_v\leftarrow(m_v\bar{\mathbf r}_v+\mathbf r_k(\ell))/(m_v+1)$.
The matching uses pre-write nodes, and the resolved node identities determine
the endpoints of the updated edges. The description history and visual
history are distinct from the relational running mean. A node can therefore
exist before a description or grounded crop has been obtained.

\paragraph{Reading in the reactive mode.}
When a question arrives, the nodes with at least one description are ranked
with \mref{eq:rank}. Up to $k{=}8$ nodes are serialized with their first
and last observation times, their first description and their two latest
descriptions. The event-log selection targets all lines when there are at most 60;
otherwise it targets the 57 most relevant and the three latest lines.
The selected lines are ordered in time. These are upper limits on the
retrieval allocation, not guarantees of the final serialized counts
(e.g., \cref{fig:appx1} contains 58 lines). The text and crop allocation
precedes the frame allocation. The latest available crops of up to $n_c{=}2$
highest-ranked nodes come next, and \mref{eq:nframes} gives the number of
frames, with $n_{\max}{=}18$. Up to $W{=}8$ of these frames are the most
recent consecutive seconds. The remaining slots hold key frames from the
times at which the retrieved nodes were last observed or described, at
least 2\,s apart. Duplicate frames are removed. The reader receives the
crops, then the frames in temporal order, then one text turn with the
memory and the question, and it answers in a single call. Relation edges
are not serialized (\mref{sec:reader}).

\paragraph{Historical frame access.}
The graph stores crops, text and timestamps rather than complete past
frames. When frame slots remain, historical key frames are retrieved from
the source video at previously observed timestamps $t\leq T$, as described
in \mref{sec:reader}. The effect of key-frame retrieval is evaluated in
\cref{tab:dev_ablation}.

\paragraph{Serialization.}
A retrieved node is written as
\emph{label (first seen $a$\,s, last seen $b$\,s): at $t_1$\,s: $d_1$; at $t_2$\,s: $d_2$; at $t_3$\,s: $d_3$},
and an event line as \emph{$t$\,s: caption}. The text turn ends with the
question and, for multiple-choice tasks, its options and an instruction to
answer with a letter.

\paragraph{Token accounting.}
The reference allocation budget is $B{=}3{,}510$ tokens, estimated from a
recent-frame window of $N{=}18$ frames. A frame is estimated at
$\kappa_f{\approx}195$ tokens and a crop at $\kappa_c{\approx}216$.
Both methods use the same reference allocation, with frame selection
following \mref{eq:nframes} and input-count checking as described in
\mref{sec:reader}. The measured reader-input costs are reported separately
in \mref{tab:efficiency}: both methods average 3.4K tokens on the common
question set, with 95th percentiles of 3.6K for \method and 3.7K for recent
frames. These measurements establish comparable input costs, rather than
an empirical maximum of 3,510 tokens. For example, the question in
\cref{fig:appx4} uses 3,469 tokens: 10 frames (1,994), 2 crops (432),
971 tokens of memory text and a 72-token prompt. Eight frames are recent
and two are historical key frames.

\paragraph{Proactive mode.}
For a proactive instruction, the writer runs every second on the current
frame, with the instruction in its prompt, and describes the objects most
relevant to the instruction first. Each node retains its first and most recent grounded observations (up to three recent crops); the serialized text contains
its first and two latest descriptions. Node ranking uses writer input
embeddings in this mode rather than the reactive CLIP text encoder.
In the per-second trigger loop, the reader receives up to $k{=}4$ nodes, up to eight
relations observed in the last 10\,s, a short summary of newly observed or
recurring relations, up to $n_c{=}2$ crops and $W{=}16$ recent frames. It is asked
``\emph{$\langle$instruction$\rangle$ Is it the right time to output ``X''?
$\langle$memory and relation changes$\rangle$ You can only answer yes or
no.}'' The first \emph{yes} triggers a second call that writes the
response. The budget $B$ is not enforced in this mode (\mref{sec:proactive}).
The first-positive rule describes a single response trigger, as in
StreamingBench PO. OmniPro Probe instead evaluates decisions around each
annotated trigger, and InteractionBench evaluates a sequence of timestamped
responses, including redundant or premature emissions
(\cref{supp:benchmarks}).

\paragraph{Prompt interfaces.}
The following specifies the inputs and outputs of the prompt stages; it
is a schematic description of their roles, not a verbatim transcript.
\begin{itemize}
\setlength{\itemsep}{0pt}
\setlength{\parsep}{0pt}
\item \textbf{Reactive writer:} current causal clip $C_k$; output scene
  triplets $(s,p,o)$, one event caption $e_k$, and short label--description
  pairs $(\ell,d)$ as in \mref{eq:writer}. No question is supplied.
\item \textbf{Reactive reader:} selected crops, frames, serialized
  object histories and event lines, then the question and answer options;
  output the answer letter for multiple-choice questions.
\item \textbf{Proactive writer:} the current frame and instruction;
  extract the same memory components while prioritizing relevant objects.
\item \textbf{Proactive trigger:} instruction, retrieved memory, relation
  changes and visual context; output a yes/no decision using the template
  given above.
\item \textbf{Proactive response:} on a positive decision, generate the
  requested response in a separate reader call.
\end{itemize}
The verbatim prompts will accompany the code.

\begin{table}[t]
\centering
\caption{\textbf{Settings of the two \method modes}, with the symbols of
\mref{sec:method}.}
\label{tab:hparams}
\scriptsize
\setlength{\tabcolsep}{2pt}
\begin{tabular}{@{}>{\raggedright\arraybackslash}p{0.29\linewidth}>{\raggedright\arraybackslash}p{0.33\linewidth}>{\raggedright\arraybackslash}p{0.33\linewidth}@{}}
\toprule
 & Reactive (question-agnostic) & Proactive (query-conditioned) \\
\midrule
Write interval $\Delta$ & 5\,s (videos $\le$300\,s), else 10\,s & 1\,s \\
Clip $C_k$ & last 5 frames at 1\,fps & 1 frame \\
Question in writer prompt & no & yes \\
Descriptions per write & $\le$3, $\le$20 words, refresh $\ge$10\,s & $\le$3, $\le$48 tokens, refresh $\ge$10\,s \\
Description order & never described, then oldest & relevance to instruction \\
Detector threshold & 0.15 & 0.15 \\
Crop & 224\,px, 50\% context & 224\,px, 50\% context \\
Crops stored per node & all & first + latest 3 \\
Match threshold $\theta$, \mref{eq:resolve} & 0.55 & 0.55 \\
Node ranking & CLIP ViT-L/14 text, \mref{eq:rank} & writer input embeddings \\
Nodes read $k$ & 8 & 4 \\
Descriptions per node & first + 2 latest & first + 2 latest \\
Relations read & none & last 10\,s, $\le$8 \\
Event lines & all if $\le$60, else 57 most relevant + 3 latest & none (relation-change summary) \\
Crops read $n_c$ & 2 & 2 \\
Recent window $W$ & 8 & 16 \\
Max.\ frames $n_{\max}$ & 18 & 16 \\
Key-frame spacing & $\ge$2\,s & -- \\
Reference budget $B$ & 3,510 & not enforced \\
\bottomrule
\end{tabular}
\end{table}

\begin{table*}[t]
\centering
\caption{\textbf{Per-task results on OVO-Bench} (accuracy, \%).
Real-time tasks: OCR; ACR, action recognition; ATR, attribute recognition;
STU, spatial understanding; FPD, future prediction; OJR, object recognition.
Backward tracing: EPM, episodic memory; ASI, action sequence identification;
HLD, hallucination detection. Tok./q, RT, BW and Avg.\ are the values of
\mref{tab:main}. A dash indicates that the corresponding task-level result is not reported.}
\label{tab:supp_ovo_pertask}
\scriptsize
\setlength{\tabcolsep}{4pt}
\renewcommand{\arraystretch}{1.0}
\resizebox{\textwidth}{!}{%
\begin{tabular}{@{}l c cccccc c ccc c c@{}}
\toprule
 & & \multicolumn{7}{c}{Real-time} & \multicolumn{4}{c}{Backward tracing} & \\
\cmidrule(lr){3-9}\cmidrule(lr){10-13}
Method & Tok./q & OCR & ACR & ATR & STU & FPD & OJR & RT & EPM & ASI & HLD & BW & Avg. \\
\midrule
\multicolumn{14}{@{}l}{\textit{InternVL3.5-8B}} \\
Recent frames & 3.4K & -- & -- & -- & -- & -- & -- & 73.9 & -- & -- & -- & 48.6 & 61.3 \\
+ \rg & 29.9K & 83.8 & 68.5 & 75.3 & 53.4 & 70.0 & 64.9 & 69.3 & 46.9 & 59.3 & 28.6 & 44.9 & 57.1 \\
\rowcolor{oursyellow}
\textbf{+ \method{} (ours)} & 3.3K & 89.5 & 72.6 & 79.2 & 60.3 & 74.4 & 75.4 & 75.2 & 55.3 & 59.3 & 57.9 & 57.5 & 66.4 \\
\midrule
\multicolumn{14}{@{}l}{\textit{LLaVA-Video-7B}} \\
Recent frames & 3.2K & -- & -- & -- & -- & -- & -- & 72.6 & -- & -- & -- & 41.7 & 57.2 \\
+ \rg & 23.9K & 77.2 & 63.8 & 78.9 & 50.8 & 70.7 & 70.2 & 68.6 & 57.4 & 53.8 & 5.1 & 38.8 & 53.7 \\
\rowcolor{oursyellow}
\textbf{+ \method{} (ours)} & 3.3K & 80.0 & 74.0 & 79.2 & 61.1 & 74.4 & 72.4 & 73.5 & 56.6 & 59.3 & 30.1 & 48.7 & 61.1 \\
\midrule
\multicolumn{14}{@{}l}{\textit{LLaVA-OV-1.5-8B}} \\
Recent frames & 3.3K & -- & -- & -- & -- & -- & -- & 74.4 & -- & -- & -- & 53.5 & 64.0 \\
\rowcolor{oursyellow}
\textbf{+ \method{} (ours)} & 3.3K & 89.5 & 76.7 & 85.7 & 60.3 & 72.2 & 76.9 & 76.9 & 50.9 & 63.0 & 76.7 & 63.5 & 70.2 \\
\midrule
\multicolumn{14}{@{}l}{\textit{Qwen3-VL-8B}} \\
Recent frames & 3.3K & -- & -- & -- & -- & -- & -- & 78.0 & -- & -- & -- & 53.9 & 66.0 \\
+ \rg & 37.5K & 90.6 & 74.3 & 75.9 & 59.6 & 69.3 & 71.7 & 73.6 & 55.6 & 66.9 & 33.9 & 52.1 & 62.9 \\
+ OASIS & 10.5K & 93.3 & 83.6 & 87.0 & 67.9 & 70.0 & 77.6 & 79.9 & 58.4 & 63.9 & 55.6 & 59.3 & 69.6 \\
\rowcolor{oursyellow}
\textbf{+ \method{} (ours)} & 3.3K & 93.3 & 78.1 & 83.1 & 65.6 & 75.6 & 78.4 & 79.0 & 57.1 & 73.1 & 53.4 & 61.2 & 70.1 \\
\midrule
\multicolumn{14}{@{}l}{\textit{Qwen3-VL-32B}} \\
Recent frames & 3.3K & -- & -- & -- & -- & -- & -- & 77.4 & -- & -- & -- & 65.2 & 71.3 \\
+ \rg & 37.5K & 93.3 & 78.1 & 85.7 & 64.9 & 75.6 & 80.6 & 79.7 & 64.2 & 71.3 & 48.7 & 61.4 & 70.6 \\
+ OASIS & 15.3K & 96.2 & 89.5 & 85.0 & 66.7 & 78.3 & 76.5 & 82.0 & 57.9 & 77.8 & 52.9 & 62.9 & 72.5 \\
\rowcolor{oursyellow}
\textbf{+ \method{} (ours)} & 3.3K & 90.5 & 75.3 & 84.4 & 67.2 & 77.8 & 80.6 & 79.3 & 63.3 & 75.0 & 72.2 & 70.2 & 74.8 \\
\midrule
\multicolumn{14}{@{}l}{\textit{Qwen3.5-9B}} \\
Recent frames & 3.3K & -- & -- & -- & -- & -- & -- & 79.8 & -- & -- & -- & 49.2 & 64.5 \\
+ \rg & 37.6K & 88.6 & 68.5 & 92.2 & 61.8 & 76.7 & 77.6 & 77.6 & 65.0 & 62.0 & 21.8 & 49.6 & 63.6 \\
+ OASIS & 17.2K & 89.5 & 74.0 & 88.3 & 61.1 & 74.4 & 76.1 & 77.2 & 60.6 & 62.0 & 30.1 & 50.9 & 64.1 \\
\rowcolor{oursyellow}
\textbf{+ \method{} (ours)} & 3.3K & 89.5 & 75.3 & 88.3 & 66.4 & 83.3 & 78.4 & 80.2 & 62.4 & 69.4 & 42.1 & 58.0 & 69.1 \\
\midrule
\multicolumn{14}{@{}l}{\textit{Qwen3.5-27B}} \\
Recent frames & 3.3K & -- & -- & -- & -- & -- & -- & 80.8 & -- & -- & -- & 58.7 & 69.8 \\
+ \rg & 37.5K & 93.3 & 80.8 & 93.5 & 67.9 & 77.8 & 87.3 & 83.4 & 66.4 & 75.0 & 36.1 & 59.2 & 71.3 \\
+ OASIS & 13.1K & 94.3 & 82.2 & 85.7 & 67.9 & 78.9 & 81.3 & 81.7 & 65.9 & 70.4 & 60.9 & 65.7 & 73.7 \\
\rowcolor{oursyellow}
\textbf{+ \method{} (ours)} & 3.3K & 96.2 & 89.5 & 80.0 & 78.8 & 82.6 & 76.5 & 83.9 & 59.6 & 85.2 & 58.8 & 67.9 & 75.9 \\
\bottomrule
\end{tabular}}
\end{table*}

\begin{table*}[t]
\centering
\caption{\textbf{Per-task results on StreamingBench} (accuracy, \%).
Real-time categories: OP, object perception; CR, causal reasoning; CS, clips
summarization; ATP, attribute perception; EU, event understanding; TR,
text-rich understanding; PR, prospective reasoning; SU, spatial
understanding; ACP, action perception; CT, counting. Tok./q, RT, PO and
Overall are the values of \mref{tab:main}. RT is the accuracy over all
real-time questions, and Overall is the mean of the ten categories and PO.
A dash indicates that the corresponding result is not reported.}
\label{tab:supp_sb_pertask}
\scriptsize
\setlength{\tabcolsep}{4pt}
\renewcommand{\arraystretch}{1.0}
\resizebox{\textwidth}{!}{%
\begin{tabular}{@{}l c cccccccccc c c c@{}}
\toprule
 & & \multicolumn{11}{c}{Real-time} & & \\
\cmidrule(lr){3-13}
Method & Tok./q & OP & CR & CS & ATP & EU & TR & PR & SU & ACP & CT & RT & PO & Overall \\
\midrule
\multicolumn{15}{@{}l}{\textit{InternVL3.5-8B}} \\
Recent frames & 3.3K & -- & -- & -- & -- & -- & -- & -- & -- & -- & -- & 79.4 & 36.8 & 74.1 \\
+ \rg & 30.9K & 78.8 & 78.9 & 82.8 & 84.2 & 81.1 & 79.3 & 75.3 & 66.5 & 70.6 & 55.5 & 75.7 & 30.0 & 71.2 \\
\rowcolor{oursyellow}
\textbf{+ \method{} (ours)} & 3.3K & 86.1 & 71.1 & 88.7 & 87.6 & 83.5 & 81.7 & 80.5 & 72.0 & 81.4 & 51.8 & 80.0 & 37.4 & 74.7 \\
\midrule
\multicolumn{15}{@{}l}{\textit{LLaVA-Video-7B}} \\
Recent frames & 3.1K & -- & -- & -- & -- & -- & -- & -- & -- & -- & -- & 78.4 & 19.5 & 71.8 \\
+ \rg & 24.4K & 79.8 & 82.7 & 82.1 & 83.3 & 78.2 & 76.9 & 80.0 & 64.1 & 77.2 & 51.4 & 75.8 & 25.3 & 71.0 \\
\rowcolor{oursyellow}
\textbf{+ \method{} (ours)} & 3.2K & 83.9 & 76.7 & 91.2 & 88.4 & 85.8 & 78.9 & 79.2 & 70.3 & 80.7 & 43.3 & 79.1 & 26.8 & 73.2 \\
\midrule
\multicolumn{15}{@{}l}{\textit{LLaVA-OV-1.5-8B}} \\
Recent frames & 3.3K & -- & -- & -- & -- & -- & -- & -- & -- & -- & -- & 77.7 & 34.7 & 72.4 \\
\rowcolor{oursyellow}
\textbf{+ \method{} (ours)} & 3.3K & 82.4 & 70.0 & 87.0 & 85.9 & 81.1 & 81.7 & 80.5 & 74.7 & 77.0 & 53.7 & 78.6 & 36.8 & 73.7 \\
\midrule
\multicolumn{15}{@{}l}{\textit{Qwen3-VL-8B}} \\
Recent frames & 3.2K & -- & -- & -- & -- & -- & -- & -- & -- & -- & -- & 79.5 & 39.5 & 74.5 \\
+ \rg & 29.4K & 84.0 & 78.1 & 88.0 & 84.6 & 74.8 & 83.5 & 83.3 & 63.0 & 69.3 & 58.0 & 77.5 & 45.8 & 73.9 \\
+ OASIS & 9.5K & 87.2 & 70.0 & 90.8 & 88.0 & 82.7 & 85.4 & 76.6 & 75.3 & 76.6 & 50.6 & 80.2 & -- & -- \\
\rowcolor{oursyellow}
\textbf{+ \method{} (ours)} & 3.2K & 88.3 & 73.3 & 95.8 & 88.8 & 83.5 & 88.2 & 88.3 & 74.2 & 81.0 & 55.5 & 83.1 & 46.3 & 78.5 \\
\midrule
\multicolumn{15}{@{}l}{\textit{Qwen3-VL-32B}} \\
Recent frames & 3.2K & -- & -- & -- & -- & -- & -- & -- & -- & -- & -- & 82.6 & 44.2 & 78.2 \\
+ \rg & 29.4K & 86.8 & 86.7 & 89.1 & 88.4 & 78.0 & 90.2 & 87.0 & 77.5 & 81.4 & 51.2 & 82.4 & 47.9 & 78.6 \\
+ OASIS & 13.2K & 87.5 & 74.1 & 94.9 & 85.7 & 81.8 & 90.9 & 85.7 & 75.9 & 78.9 & 59.4 & 82.9 & -- & -- \\
\rowcolor{oursyellow}
\textbf{+ \method{} (ours)} & 3.2K & 87.5 & 83.3 & 96.2 & 90.5 & 87.4 & 88.6 & 87.0 & 74.7 & 79.6 & 53.0 & 83.6 & 48.4 & 79.7 \\
\midrule
\multicolumn{15}{@{}l}{\textit{Qwen3.5-9B}} \\
Recent frames & 3.2K & -- & -- & -- & -- & -- & -- & -- & -- & -- & -- & 84.3 & 35.0 & 79.0 \\
+ \rg & 29.4K & 90.1 & 87.8 & 92.4 & 90.0 & 81.9 & 89.0 & 85.7 & 78.6 & 82.5 & 54.9 & 84.2 & 43.5 & 79.7 \\
+ OASIS & 17.1K & 87.5 & 71.1 & 88.2 & 88.4 & 85.8 & 83.7 & 71.4 & 69.8 & 77.3 & 52.4 & 79.5 & -- & -- \\
\rowcolor{oursyellow}
\textbf{+ \method{} (ours)} & 3.2K & 89.7 & 87.8 & 95.0 & 92.1 & 88.2 & 90.2 & 87.0 & 76.9 & 83.3 & 50.6 & 85.0 & 47.0 & 80.7 \\
\midrule
\multicolumn{15}{@{}l}{\textit{Qwen3.5-27B}} \\
Recent frames & 3.2K & -- & -- & -- & -- & -- & -- & -- & -- & -- & -- & 85.9 & 43.0 & 81.1 \\
+ \rg & 29.4K & 90.8 & 92.2 & 92.9 & 90.5 & 85.0 & 89.8 & 84.4 & 76.4 & 84.8 & 62.8 & 85.7 & 38.9 & 80.8 \\
+ OASIS & 10.8K & 87.5 & 88.9 & 93.2 & 90.5 & 78.8 & 92.4 & 100.0 & 77.6 & 77.5 & 53.1 & 84.0 & -- & -- \\
\rowcolor{oursyellow}
\textbf{+ \method{} (ours)} & 3.2K & 91.6 & 92.2 & 94.5 & 92.1 & 88.2 & 92.3 & 87.0 & 79.7 & 85.5 & 56.7 & 86.7 & 54.7 & 83.1 \\
\bottomrule
\end{tabular}}
\end{table*}

\section{Benchmark details}
\label{supp:benchmarks}

\paragraph{OVO-Bench~\cite{li2025ovobench}}
has 12 tasks over 644 videos and asks each question at a given time in the
stream. Its six real-time tasks concern the current scene: OCR, action
recognition (ACR), attribute recognition (ATR), spatial understanding (STU),
future prediction (FPD) and object recognition (OJR). Its three
backward-tracing tasks concern earlier parts of the stream: episodic memory
(EPM), action sequence identification (ASI) and hallucination detection
(HLD), which asks about content that the video does not show. Both groups
are multiple choice. As in the benchmark, the real-time and backward scores
of \mref{tab:main} are unweighted means over their tasks, and Avg.\ is the
mean of the two. The benchmark also has three forward active responding
tasks. We do not report them in \mref{tab:main}, and we evaluate proactive
behavior on StreamingBench, OmniPro and InteractionBench instead.

\paragraph{StreamingBench~\cite{lin2024streamingbench}.}
We use its real-time visual understanding set, which has 2,500
multiple-choice questions over 500 videos in ten categories: object
perception (OP), causal reasoning (CR), clips summarization (CS), attribute
perception (ATP), event understanding (EU), text-rich understanding (TR),
prospective reasoning (PR), spatial understanding (SU), action perception
(ACP) and counting (CT). We also use its proactive output (PO) task, in
which the model must emit a given output once a condition occurs and is
correct only if it does so within 2\,s of the annotated time. The real-time
score is the accuracy over all real-time questions, whereas Overall in
\mref{tab:main} is the mean of the ten category accuracies and PO. PO is
answered with the query-conditioned memory.

\paragraph{OmniPro~\cite{omnipro}}
has 2,700 samples over 1,262 videos in nine proactive sub-tasks: Event Alert
(alert when a given event occurs), Target Grounding (report where a target
is when it appears), State Monitoring (report each change of a monitored
state), Snapshot Counting (count the targets visible at each trigger),
Conditional Alert (alert when a semantically defined condition holds),
Cumulative Counting (keep a running count of a repeated event), Event
Narration (narrate the stream at natural breakpoints), Deduplicated Counting
(count distinct targets, ignoring re-appearances) and Step Instruction (give
the next step of a tutorial). \Mref{tab:further} reports the accuracy of its
Probe mode, which queries the model shortly before and shortly after each
annotated trigger. A trigger is correct only if the model does not respond
before it and answers correctly after it. Mean is the unweighted mean of the
nine sub-tasks. Probe mode supplies the evaluation times; unlike the
benchmark's Online mode, it does not require the system to discover all
trigger times autonomously.

\paragraph{InteractionBench~\cite{interactionbench}}
has 1,060 interactions over 812 videos and scores a system on the clock of
the video. Content accuracy (Acc.) grades what a matched response says.
Timing accuracy (TA) falls with the delay of the matched response, and
silence compliance (SC) falls with the number of redundant and premature
responses. Overall is computed per evaluation item before aggregation,
following the benchmark's item-specific scoring rules. For question,
trigger and counting items, the score averages content accuracy and the
harmonic mean of TA and SC (using TA alone when SC is undefined).
Commentary and belief-revision items use content accuracy, while
silence-only items use SC. Overall is the mean of the item scores; each
reported component is averaged over items for which it is defined.
Consequently, Overall cannot in general be reconstructed from the three
aggregate columns in \mref{tab:further}.

\paragraph{Splits.}
We set aside a development split of 1,088 questions and used it only to
choose the read configuration of \cref{tab:hparams}
(\cref{sec:dev_ablation}). All reader-matched results in \mref{tab:main}, including recent frames,
Response-G1, OASIS and \method, use identical evaluation questions for each
reported task. The development PO subset contains 45 questions.

\paragraph{Published results.}
The top three blocks of \mref{tab:main} and the cited gray rows of
\mref{tab:main,tab:further} are published results. They follow their own
input protocols and use the full question sets, so we list them for context
and do not compare them with the reader-matched rows at equal input.

\section{Per-task results}
\label{supp:pertask}

\Cref{tab:supp_ovo_pertask,tab:supp_sb_pertask} report the available
per-task results for recent frames, \rg~\cite{ma2026responseg1},
OASIS~\cite{liang2026oasis} and \method. The aggregate columns repeat
\mref{tab:main}; task-level results that are not reported are marked with a dash.
On OVO-Bench, backward tracing improves over recent frames for all seven
readers, by 8.9, 7.0, 10.0, 7.3, 5.0, 8.8 and 9.2 points in the order of the
table. On StreamingBench, real-time accuracy improves for all seven readers,
by 0.6, 0.7, 0.9, 3.6, 1.0, 0.7 and 0.8 points.

\section{Additional ablations}
\label{sec:supp_ablation}

\Mref{sec:ablation} discusses the first four rows of \mref{tab:ablation}.
Here we discuss the association and state-history rows and two further ablations,
appearance-only matching and swapped object histories.
\Cref{tab:supp_ablation} lists all of them. Every ablation uses the frozen
Qwen3-VL-8B reader and changes one property of the memory or of the reader
input. These changes can also affect which evidence fits within the nominal
budget, so the results evaluate complete reader configurations rather than
holding the exact retrieved tokens fixed.

\begin{table}[t]
\centering
\caption{\textbf{Ablations with Qwen3-VL-8B} (accuracy, \%). The unmarked
rows repeat \mref{tab:ablation}, and the two rows marked $\dagger$ are new.
RT is real-time accuracy and BW is backward tracing.}
\label{tab:supp_ablation}
\scriptsize
\setlength{\tabcolsep}{3.5pt}
\renewcommand{\arraystretch}{1}
\begin{tabular}{@{}lccc@{}}
\toprule
 & \textbf{StreamingBench} & \multicolumn{2}{c}{\textbf{OVO-Bench}}\\
\cmidrule(lr){3-4}
\textbf{Configuration} & RT & RT & BW\\
\midrule
Recent frames only & 79.5 & 78.0 & 53.9\\
Memory only (no recent frames) & 68.1 & 62.6 & 54.8\\
No object nodes & 79.8 & 77.8 & 57.5\\
No consolidation (flat list) & 79.5 & 75.6 & 59.5\\
\midrule
Label-only matching & 78.4 & 77.5 & 59.8\\
Appearance-only matching$^{\dagger}$ & 81.2 & 77.2 & 58.2\\
\midrule
Timestamps shuffled & 79.0 & 76.3 & 59.0\\
Latest description only & 78.8 & 77.4 & 60.6\\
Swapped object histories$^{\dagger}$ & 76.8 & 73.5 & 52.0\\
\midrule
\textbf{Full \method} & \textbf{83.1} & \textbf{79.0} & \textbf{61.2}\\
\bottomrule
\end{tabular}
\end{table}

\paragraph{Relational association.}
Two ablations change how a mention is matched to a stored node.
\emph{Label-only matching} merges a mention into the stored node of the same
label class and ignores its relational context. \emph{Appearance-only
matching} compares the crops of the objects instead of their relational
contexts. Both lower all three scores. Label-only matching loses 4.7 points
of real-time accuracy on StreamingBench and 1.4 points of backward tracing,
and appearance-only matching loses 1.9 and 3.0 points. Matching by the
relations an object takes part in gives higher downstream accuracy than
the tested label- and appearance-only alternatives.

\paragraph{Object-state histories.}
Three ablations change the history that a node carries. \emph{Timestamps
shuffled} permutes the times of the stored descriptions, and \emph{latest
description only} keeps a single description per node. They lower backward
tracing by 2.2 and 0.6 points, and real-time accuracy on StreamingBench by
4.1 and 4.3 points. \emph{Swapped object histories} exchanges the
description histories of the nodes, so each object is read with the history
of another object. This is the largest degradation among the history
ablations and the lowest backward-tracing score in the table. All three
scores fall below those of recent frames alone (76.8 against 79.5, 73.5
against 78.0, and 52.0 against 53.9). This highlights the importance of maintaining consistent associations between persistent objects and their timestamped state histories.

\section{Development-split ablations}
\label{sec:dev_ablation}

\Cref{tab:dev_ablation} lists the one-change study on the development split
(\cref{supp:benchmarks}) that we used to choose the read configuration of
the reactive mode. Its macro average covers five task families, including
the forward tasks of OVO-Bench and PO, both answered here with the
question-agnostic memory. The development PO subset contains 45 questions. Removing the object nodes costs the most in the
development macro score ($-4.6$ points), and backward tracing drops from
58.4 to 50.0. This provides evidence complementary to the held-out no-object
ablation in \cref{tab:supp_ablation}. Removing historical key frames changes
backward tracing from 58.4 to 55.6 and the macro score from 59.2 to 57.9,
indicating that historical images contribute but are not necessary to run
the memory-based reader. Serializing relation edges does not improve the
macro score, which supports retaining edges for consolidation rather than
passing them directly to the reactive reader (\mref{sec:reader}).

\begin{table}[t]
\centering
\caption{\textbf{Design-time ablations} on the development split (1,088
questions, Qwen3-VL-8B, $B{=}3{,}510$; accuracy, \%). RTVU is StreamingBench
real-time; RT, BW and FW are OVO-Bench real-time, backward tracing and
forward active responding; PO is StreamingBench proactive output. FW and
PO use question-agnostic memory in this study, unlike the final proactive
configuration. Macro averages the five columns; $\Delta$ is the difference
between the displayed macro scores and the selected configuration.}
\label{tab:dev_ablation}
\setlength{\tabcolsep}{3pt}
\small
\resizebox{\columnwidth}{!}{%
\begin{tabular}{@{}lccccccc@{}}
\toprule
Variant & RTVU & RT & BW & FW & PO & Macro & $\Delta$ \\
\midrule
\textbf{Selected configuration} & 82.9 & 71.0 & \textbf{58.4} & 46.1 & \textbf{37.8} & \textbf{59.2} & -- \\
Event log only (no object nodes) & 82.4 & 70.7 & 50.0 & 45.5 & 24.4 & 54.6 & $-4.6$ \\
No crops & 82.5 & 71.5 & 54.7 & 46.4 & 26.7 & 56.4 & $-2.8$ \\
Key frames at event times & 81.9 & 70.8 & 53.0 & 46.0 & 33.3 & 57.0 & $-2.2$ \\
4 nodes instead of 8 & 82.7 & 69.7 & 56.4 & 46.9 & 31.1 & 57.4 & $-1.8$ \\
Most recent nodes & 83.2 & 70.9 & 55.9 & 46.8 & 31.1 & 57.6 & $-1.6$ \\
15 event lines instead of 60 & 82.2 & 71.9 & 56.1 & 47.1 & 31.1 & 57.7 & $-1.5$ \\
No event log & 82.5 & 70.4 & 56.9 & 46.1 & 33.3 & 57.9 & $-1.3$ \\
No key frames & 82.4 & 71.9 & 55.6 & 44.1 & 35.6 & 57.9 & $-1.3$ \\
$+$ relations & 82.4 & 71.6 & 55.6 & 47.2 & 35.6 & 58.5 & $-0.7$ \\
\bottomrule
\end{tabular}}
\end{table}

\section{Efficiency details}
\label{supp:efficiency}

\Mref{sec:efficiency} compares three quantities that should be kept apart:
the input that the reader receives for a question, the memory that a method
has stored, and the cost of writing that memory.

\paragraph{Columns of \mref{tab:efficiency}.}
Tokens are reader input tokens per question, as mean and 95th percentile. KV is the mean key-value cache
memory of the reader, and Calls/q is the mean number of times the VLM is run
to answer one question. Memory is the median number of tokens that a method
has stored after 24 minutes of stream, which is not what it reads for a
question. Stored-token counts quantify the representation in token units;
they should not be equated with physical storage in bytes or reader KV-cache
size.

\begin{figure}[t]
\centering
\includegraphics[width=\columnwidth]{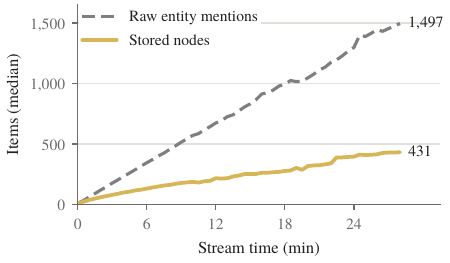}
\caption{\textbf{Consolidation over time.} Raw entity mentions and stored
nodes as the stream grows (median over the videos of at least 10 minutes).
After 28 minutes, 1,497 mentions are held in 431 nodes.}
\label{fig:qualstats}
\end{figure}

\paragraph{Memory representation and consolidation.}
The reactive allocation targets the recent-frame input cost using the
nominal budget $B$ (\cref{supp:impl}). The stored memory grows with the stream, independently of the reader allocation. Consolidation reduces the number
of persistent nodes relative to unmerged mentions. In the stream of
\mref{fig:overview}, 362 entity mentions become 98 nodes and 40 event lines
after 409\,s, and the longest example (\cref{fig:appx1}) holds 138 nodes
at a question time of 658\,s. The memory keeps crops and text, and no
complete frames. The 59K stored-token result in \mref{tab:efficiency}
quantifies the VSG representation itself; this representation-level measurement excludes the source video.

\paragraph{Reading latency.}
\Mref{tab:efficiency} measures reactive answering on the common question
set with Qwen3-VL-8B on an L40S. \method and recent frames use Hugging Face
Transformers; OASIS uses its official Transformers implementation, and
\rg uses vLLM. Both \method and recent frames average 3.4K reader input
tokens; the reported 1.0\,s versus 5.2\,s latency difference measures the reactive answering pipeline.
As explained in \mref{sec:efficiency}, the reader times are similar and
fewer frame decodes account for the reported gain. This comparison excludes
the separate cost of constructing the graph described below.

\paragraph{Writing cost.}
The question-agnostic writer costs 2.06 GPU-hours per hour of video, with a
peak of 19.5\,GB of GPU memory. It runs once per video, and all questions on
that video and all seven readers reuse its output. The proactive mode writes
one memory per instruction. Writing cost is measured separately from reactive
answering latency. As stated in \mref{sec:conclusion}, reducing the cost of
writing and storing the memory is left to future work.

\section{Qualitative analysis}
\label{supp:qual}

\begin{figure*}[!t]
\centering
\includegraphics[width=0.85\textwidth]{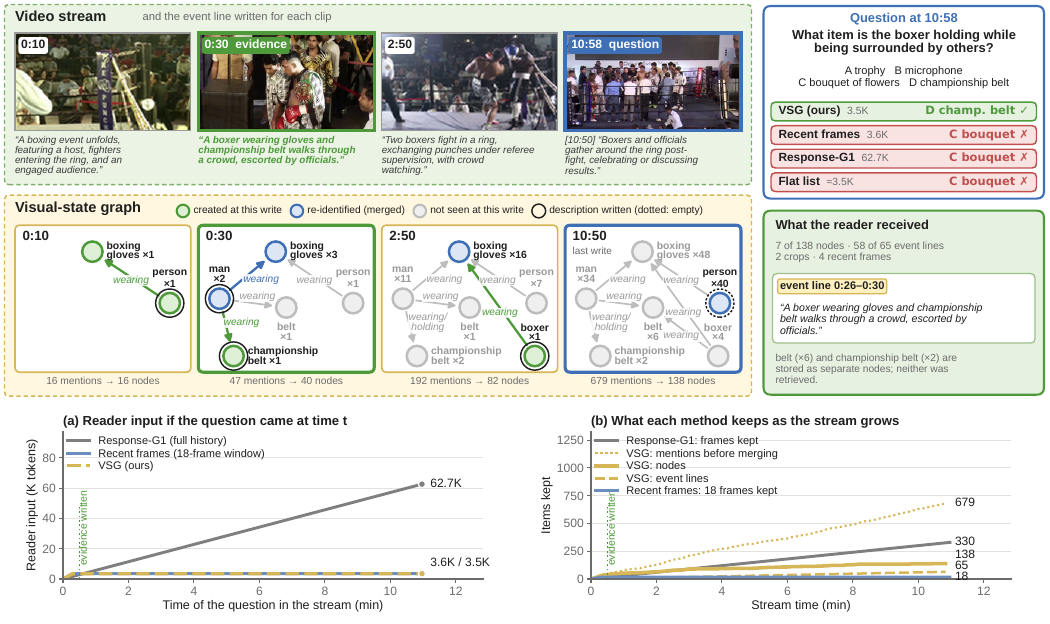}
\caption{\textbf{Example 1, OVO-Bench real-time (OJR).} The belt is shown
and described at 0:30, and the question comes at 10:58, 628\,s later.
\method answers D; the 0:30 event line contains the relevant item.
Recent frames, full-history \rg
and the flat list answer C. \emph{belt} and \emph{championship belt} are
stored as separate nodes, and neither was retrieved.}
\label{fig:appx1}
\end{figure*}

\begin{figure*}[!t]
\centering
\includegraphics[width=0.85\textwidth]{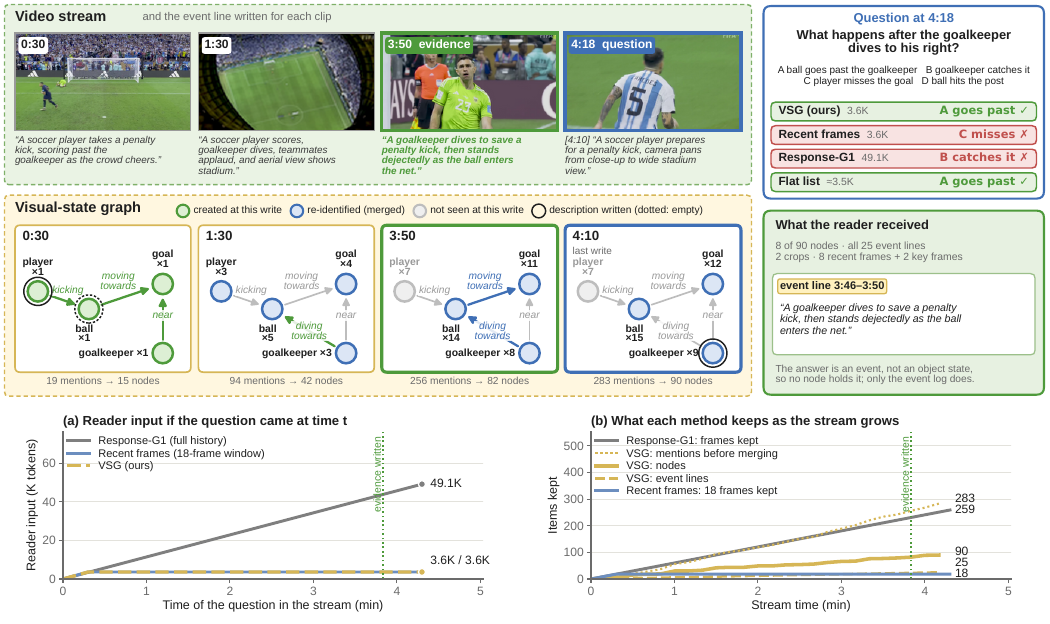}
\caption{\textbf{Example 2, StreamingBench real-time (event
understanding).} The save attempt and the goal are written at 3:50, and the
question comes at 4:18, 28\,s later and just outside the 18-s window.
\method and the flat list answer A. Recent frames answer C, and \rg answers
B. The retrieved event line explicitly records the outcome of the save
attempt.}
\label{fig:appx3}
\end{figure*}

\begin{figure*}[!t]
\centering
\includegraphics[width=0.85\textwidth]{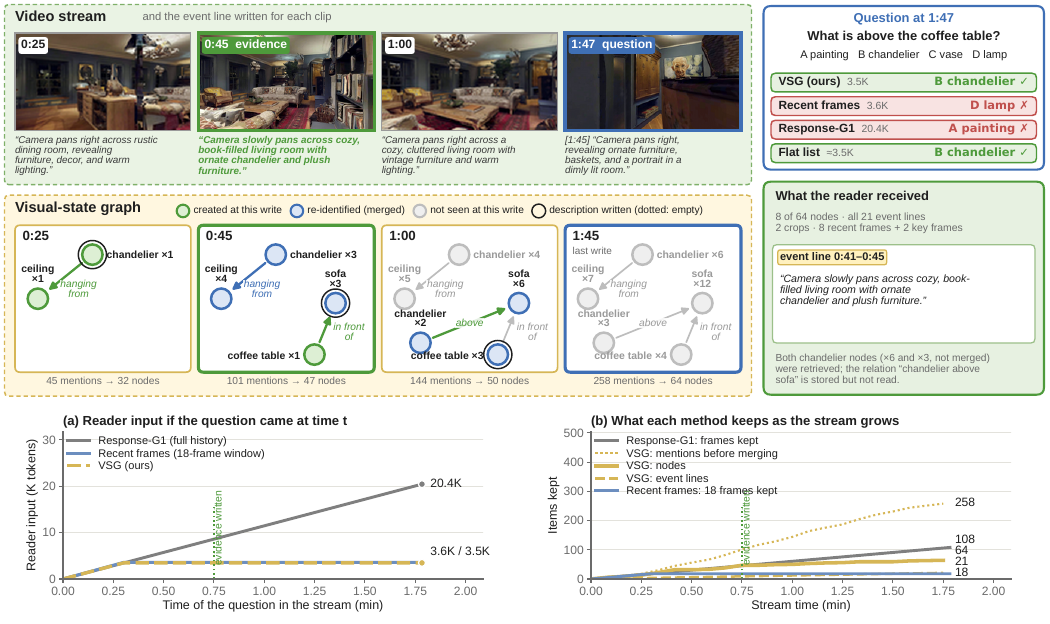}
\caption{\textbf{Example 3, OVO-Bench backward tracing (EPM).} The
chandelier is written to the event log at 0:45, and the question comes at
1:47, 62\,s later. \method and the flat list answer B. Recent frames answer
D, and \rg answers A. The two crops that the reader received show the
fireplace and the floor, not the chandelier. Both \emph{chandelier} nodes,
which were not merged, were retrieved together with the event line. The
retrieved event line names a chandelier, and the reader also receives a
chandelier description from 25\,s. The caption alone does not specify its
position relative to the coffee table.}
\label{fig:appx4}
\end{figure*}

\paragraph{Setup.}
All examples use the reactive mode with Qwen3-VL-8B as the reader. The
inputs of the reader are recomputed deterministically from the stored
memory. The frames are decoded again at the logged times, and the graphs,
descriptions and crops are the stored memory items, drawn from the write
log. The answers and
token counts of \rg come from its full-history runs. We ranked the
candidates by a fixed score before drawing any figure. We then checked by
hand the 22 highest-ranked questions that \method answers correctly and
recent frames do not, and whose supporting line is older than the recent
window (HLD excluded). We retained examples whose event line explicitly states the content of the correct option. We drew
six examples, four of them from the test split. One is the example of
\mref{fig:overview}, and the other three are shown in
\cref{fig:appx1,fig:appx3,fig:appx4}.

\paragraph{How to read the example figures.}
Each figure follows the layout of \mref{fig:overview}. \emph{Video stream.}
Four writes of the stream are shown, each with its frame and, below it, the
event line that the writer stored for that clip. The write containing
evidence for the correct answer is marked in green. The last column shows the
frame at the question time, above the last write before the question.
\emph{Visual-state graph.} Each tile shows a part of the \mg around the
objects of the question, at the same four writes. A node is drawn once it
has a relation to another node of the tile, and relations to objects
outside the tile are not drawn. Green nodes and relations
are created at that write, blue ones are matched to a stored node or
observed again, and gray ones are stored but not seen in that clip.
$\times n$ counts the observations merged into a node so far. A black ring
marks a description written at that write, and a dotted ring an empty ``not
visible'' description. The line under each tile gives the raw entity
mentions and the persistent nodes of the whole memory so far. \emph{Right.}
The question, with the answer and the measured input of each method:
\method, the 18-frame recent window, full-history \rg, and the same
descriptions and event lines as a flat list without consolidation. Below
it, what the reader of \method received, with the supporting event line.
\emph{Plots.} Plot (a) illustrates the input cost as a function of a
hypothetical question time. Only the dots at the actual question time are
measured; the curves are illustrative estimates. The recent window and
\method target the nominal budget, and the \rg curve assumes constant
cost per frame based on the measured question. Plot (b) shows what each method keeps as the stream grows: \rg
keeps every frame it samples, the recent window keeps 18 frames, and
\method keeps persistent nodes and event lines. The dotted line shows the
entity mentions that \method wrote before merging. These are counts of
items, not tokens.

\paragraph{What the examples show.}
In all four test examples, including that of \mref{fig:overview}, a
retrieved timestamped event line contains evidence relevant to the answer.
The lines were written 28 to 628\,s before the question, outside the
18-frame recent window. \rg reads the full history (20.4K
to 77.8K tokens) and answers all four wrongly. Consolidation compresses the
memory, for example from 362 mentions to 98 nodes in the stream of
\mref{fig:overview} and from 679 to 138 in \cref{fig:appx1}, and \cref{fig:qualstats} shows the same
trend over the long videos. Across these examples, the reported input
of \method is approximately 3.5--3.6K tokens. The answer objects were kept as merged nodes in
three of the four examples, but only the two \emph{chandelier} nodes of
\cref{fig:appx4} were among the nodes sent to the reader. In the example of
\mref{fig:overview}, the \emph{screwdriver} node was not retrieved, and the
two retrieved crops are associated with empty ``not visible'' descriptions.
Scene-level events such as ``the ball enters the net'' (\cref{fig:appx3})
are explicitly represented in the event log. The flat list without
consolidation answers three of the four questions and fails on
\cref{fig:appx1}. These examples illustrate historical evidence retrieval and the compact representation of repeated observations. The controlled association and history ablations in \cref{tab:supp_ablation} quantify the contribution of relational consolidation and consistent object-state histories at the aggregate level.

\clearpage

\makeatletter
\setlength{\@dblfptop}{0pt}
\setlength{\@dblfpbot}{0pt plus 1fil}
\makeatother

\end{document}